\documentclass{article}

\usepackage{arxiv}

\usepackage[utf8x]{inputenc} 
\usepackage[T1]{fontenc}    
\usepackage[pagebackref,breaklinks,colorlinks]{hyperref}       
\usepackage{url}            
\usepackage{booktabs}       
\usepackage{amsfonts}       
\usepackage{amsmath}
\usepackage{amssymb}
\usepackage{nicefrac}       
\usepackage{microtype}      
\usepackage[capitalize]{cleveref}       
\usepackage{lipsum}         
\usepackage{graphicx}
\usepackage{natbib}
\usepackage{doi}

\usepackage{float}
\usepackage{caption,setspace}
\usepackage{subcaption}
\usepackage{multicol}
\usepackage{multirow}
\usepackage{arydshln}
\usepackage{pifont}
\usepackage{listings}
\usepackage{color}
\usepackage[toc]{appendix}

\usepackage{calrsfs}
\DeclareMathAlphabet{\pazocal}{OMS}{zplm}{m}{n}
\newcommand{\Lb}{\pazocal{L}}

\crefname{section}{Sec.}{Secs.}
\Crefname{section}{Section}{Sections}
\Crefname{table}{Table}{Tables}
\crefname{table}{Tab.}{Tabs.}

\title{Boosting Semantic Segmentation by Conditioning the Backbone with Semantic Boundaries}

\date{March 16, 2023}

\author{\href{https://orcid.org/0000-0003-1494-3635}{\includegraphics[scale=0.06]{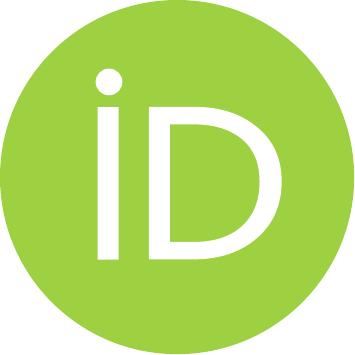}\hspace{1mm}Haruya Ishikawa}\thanks{\href{https://github.com/haruishi43}{github.com/haruishi43}} \qquad Yoshimitsu Aoki \\
	Department of Electrical Engineering \\
	Keio University\\
	Yokohama, Kanagawa 223-0061 \\
	\texttt{haruyaishikawa@keio.jp} \\
}

\renewcommand{\shorttitle}{Boosting Semantic Segmentation by Conditioning the Backbone with Semantic Boundaries}

\hypersetup{
pdftitle={Boosting Semantic Segmentation by Conditioning the Backbone with Semantic Boundaries},
pdfsubject={},
pdfauthor={Haruya Ishikawa, Yoshimitsu Aoki},
pdfkeywords={Computer Vision, Semantic Segmentation, Semantic Boundary Detection},
}

\begin{document}
\maketitle

\begin{abstract}


In this paper, we present the Semantic Boundary Conditioned Backbone (SBCB) framework, a simple yet effective training framework that is model-agnostic and boosts segmentation performance, especially around the boundaries.
Motivated by the recent development in improving semantic segmentation by incorporating boundaries as auxiliary tasks, we propose a multi-task framework that uses semantic boundary detection (SBD) as an auxiliary task.
The SBCB framework utilizes the nature of the SBD task, which is complementary to semantic segmentation, to improve the backbone of the segmentation head.
We apply an SBD head that exploits the multi-scale features from the backbone, where the model learns low-level features in the earlier stages, and high-level semantic understanding in the later stages.
This head perfectly complements the common semantic segmentation architectures where the features from the later stages are used for classification.
We can improve semantic segmentation models without additional parameters during inference by only conditioning the backbone.
Through extensive evaluations, we show the effectiveness of the SBCB framework by improving various popular segmentation heads and backbones by $0.5\%\sim3.0\%$ IoU on the Cityscapes dataset and gains $1.6\%\sim4.1\%$ in boundary Fscores.
We also apply this framework on customized backbones and the emerging vision transformer models and show the effectiveness of the SBCB framework.

\end{abstract}


\section{Introduction}
\label{sec:introduction}

Semantic segmentation is an actively studied field in computer vision and is crucial for various challenging applications such as autonomous driving and virtual reality.
Semantic segmentation is a pixel-wise classification task where each pixel represents a category.
A standard metric of quantifying segmentation quality is the intersection-over-union (IoU) metric, defined as the ratio of the intersection of the predicted segmentation mask and the ground-truth (GT) segmentation mask to the union of the two masks.
With most methods competing for the best IoU score, the boundary quality of the segmentation masks is often overlooked \cite{cheng2021boundaryiou}.
However, more precise object segmentation masks can significantly benefit various downstream applications, such as object proposal generation \cite{bertasius2015hfl}, depth estimation \cite{ramamonjisoa2020boundarydepthestimation}, and image localization \cite{ramalingam2010SKYLINE2GPSLI}.

\begin{figure}[t]
  \centering
  \includegraphics[width=0.6\linewidth]{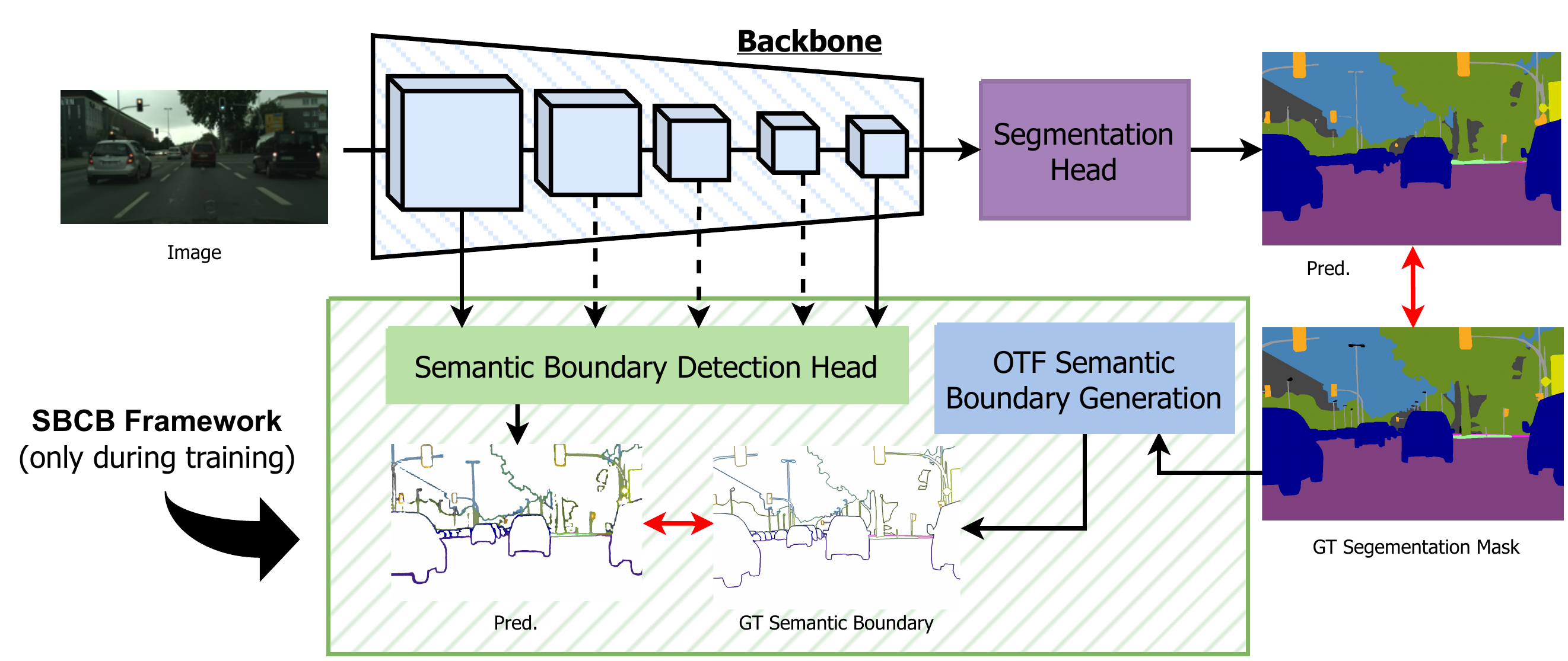}
  \caption{A simple overview of the Semantic Boundary Conditioned Backbone (SBCB) framework. The semantic boundary detection (SBD) head is applied to the backbone of the semantic segmentation head during training. The on-the-fly (OTF) semantic boundary generation module generates ground-truth (GT) semantic boundaries to train the SBD head. This simple framework improves the segmentation quality because the task of SBD is complementary but more challenging than the main task, which forces the backbone network to explicitly and jointly model boundaries and the relation to semantics.}
  \label{fig:intro_diagram}
  \vspace{-1.5em}  
\end{figure}

Closely related to semantic segmentation, semantic boundary detection (SBD) is also an active computer vision research topic.
SBD is a multi-label classification task formulation of the classical binary edge detection task, which requires the model to classify the edges and their category.
Since boundaries are always surrounding the segmentation map, SBD is often considered a dual problem of semantic segmentation.

Joint modeling of segmentation and boundary detection has recently become popular to combat the issues of poor boundary quality in semantic segmentation \cite{takikawa2019gscnn,li2020decouple,zhen2020rpcnet,yu2021coupled}.
Not only do these approaches improve segmentation accuracy around the boundaries, but they also prove that explicit modeling of the boundaries improves the overall IoU as well.
The most common approach for joint modeling is to propose a novel method of using the features learned in the boundary heads to improve the segmentation quality.
Although effective, these methods require specific architectures and are not easily transferable to other segmentation models.
The SegFix \cite{yuan2020segfix} method is a notable exception, an effective post-processing method that improves the segmentation quality by fixing the segmentation errors around the boundaries.
However, SegFix requires the user to train a separate post-processing model and adds another step during inference.
We argue that we can intrinsically improve the segmentation quality by conditioning the backbone of the segmentation head on the semantic boundaries, a technique that is model-agnostic and can be applied to any hierarchical backbones.

To this end, we present the Semantic Boundary Conditioned Backbone (SBCB) framework, a training framework aimed at boosting the segmentation quality of various segmentation architectures.
In this framework, we add a lightweight SBD head on the backbone of the segmentation network during training and perform multi-task training.
The SBD head is specifically designed so that the earlier stages of the backbones are conditioned on low-level features, and the later stages on higher-level semantic understanding.
We can discard the SBD head during inference, retaining the benefits of the conditioned backbone without any computational costs or an increase in network parameters.
The models trained using our framework consistently improve significantly in their metrics, especially around the mask boundaries.
We show effectiveness by applying our framework to various segmentation models with varying segmentation heads and backbones.
The contributions are as follows:
\begin{itemize}
  \item We propose a model-agnostic training framework aimed at conditioning the backbone for semantic segmentation called the Semantic Boundary Conditioned Backbone (SBCB) framework. This is the first training framework that utilizes semantic boundaries as an auxiliary task to improve various segmentation models both in terms of IoU and boundary Fscore. Our framework only uses the SBD head during training and does not add any computational costs during inference. We provide extensive experiments to prove the effectiveness of the framework.
  \item We propose the Binary Boundary Conditioned Backbone (BBCB) framework to compare with the SBCB framework and show that SBD is the perfect auxiliary task. The use of binary boundaries and edges has been vaguely proposed by previous works as auxiliary tasks for a specific architecture, yet it has not been made into a generalized framework compatible with various architectures.
  \item We propose applying our framework to customized architectures such as BiSeNet, STDC, and the recent vision transformers.
  \item We propose methods of utilizing the SBD head used in the SBCB framework for explicit feature fusion and show how the SBCB framework further contributes to the research in multi-task models of semantic segmentation and SBD.
  \item The SBCB framework is open-sourced to benefit the community. 
\end{itemize}

\section{Related Work}
\label{sec:related_work}

\noindent\textbf{Semantic Segmentation.}
In computer vision, semantic segmentation is one of the most popular and challenging tasks and boasts a rich set of prior works.
Long et al. \cite{shelhamer2015fcn} proposed an end-to-end trainable fully convolutional network adapted from image classification models.
In \cite{chen2017deeplabv3}, the authors introduced dilated convolution and atrous spatial pyramid pooling (ASPP) to capture multi-scale contextual information.
Zhao et al. \cite{zhao2017pspnet} proposed a pyramid pooling module (PPM) to model multi-scale contexts.
Methods introduced in \cite{fu2019dan,zhu2019ann,pang2019seenet,huang2019ccnet,fu2019danet,cao2020gcnet,yin2020dnl,fu2021dranet} achieved greater recognition of local and global context through the introduction of non-local operators \cite{wang2018nonlocal} and self-attention mechanism \cite{vaswani2017attention}.
Recently, the use of vision transformers \cite{dosovitskiy2021vit,liu2021swin} for semantic segmentation has become popularized due to its capability of learning long-range contexts \cite{Strudel2021segmenter,xie2021segformer}.
In this paper, we do not explicitly explore new methods of contextual modeling for semantic segmentation.
Instead, we introduce a framework that can be easily integrated with these models and demonstrate how our framework can improve upon these baselines.

Meanwhile, there have been works for directly modeling boundary information for segmentation using novel loss functions \cite{chen2020semeda,wang2022activeboundaryloss}.
Our work focuses on multi-task learning of semantic segmentation and boundaries, which can also incorporate these loss functions.

\noindent\textbf{Edge and Semantic Boundary Detection.}
Similar to semantic segmentation, edge and boundary detection have been widely studied.
Xie et al. \cite{xie2015hed} introduced a CNN model that can be trained end-to-end, which paved the way for various edge detection models like \cite{liu2017rcf,pu2022edter}.
Yu et al. \cite{yu2017casenet} extended the task of binary edge detection to semantic boundary detection (SBD) by formulating the problem as multi-label pixel-wise classification.
Hu et al. \cite{hu2019dff} introduced a dynamic fusion model with adaptive weights for better contextual modeling.
DDS \cite{liu2022dds} proposed a deep supervision framework that supervises all side outputs and is currently the state-of-the-art method for SBD.

\noindent\textbf{Multi-Task Learning.}
In this paper, we specify multi-task learning (MTL) as an explicit joint modeling of two or more tasks like the method introduced in \cite{misra2016crossstitch,kokkinos2017ubernet,xiao2018unified,xu2018padnet}.
While most of the models in computer vision are task-specific, there is great interest in joint modeling.
Solving multiple problems with a single model could create efficient systems and improve recognition for general AI, such as embodied agents \cite{xia2018gibson,narasimhan2020seeing}.
In MTL, it is common to use a multi-head architecture with a shared backbone for memory efficiency.
The backbone is aimed to learn a shared representation between the tasks, but often times this fails due to the backbone being designed for a single-task, leading to worse results \cite{kokkinos2017ubernet,misra2016crossstitch}.
The works of \cite{Liu2018MTAN} explores novel mechanism for obtaining features by adding task-specific attention modules in the backbone.
Our work, however, explores the use of two-head architecture, where the auxiliary semantic boundary detection task is complementary to the main segmentation task.

In semantic segmentation, edges and boundaries have been used as auxiliary tasks.
Takikawa et al. \cite{takikawa2019gscnn} introduced an MTL framework using binary boundary detection as an auxiliary task to improve semantic segmentation, especially for pixels near mask boundaries.
Similarly, Li et al. \cite{li2020decouple} introduced a novel framework for explicitly modeling the body and edge features.
This paper explores the joint modeling of semantic segmentation and semantic boundaries as an MTL framework for conditioning the backbone features.

Zhen et al. \cite{zhen2020rpcnet} introduced the first joint semantic segmentation and boundary detection (JSB) model and proposed the iterative pyramid context module and duality loss that enforces consistency between the two tasks.
Yu et al. \cite{yu2021coupled} proposed a dynamic graph propagation approach to couple the two tasks and refine segmentation and boundary maps.
In this paper, we introduce a simple yet effective modular multi-headed model that does not require complex modeling to explicitly fuse the two tasks.
We show that a shared backbone is enough to improve both tasks significantly.
We also show that we can develop a JSB model using the semantic boundary head used in our framework, which can further boost semantic segmentation performance.

\noindent\textbf{SegFix.}
SegFix \cite{yuan2020segfix} is a model-agnostic post-processing network that refines the output of a segmentation model with an independent network.
The key idea of this method is to replace unreliable predictions in the mask boundaries with reliable interior labels.
SegFix is similar to our approach in that we both aim to improve segmentation quality using boundaries in a model-agnostic way.
The key difference is that our method is a training framework, whereas SegFix requires training another model and two-step inference.
In fact, SegFix can be combined with our framework to boost performance, which we will show in this paper.


\begin{figure}[ht]
  \centering
  \includegraphics[width=0.5\linewidth]{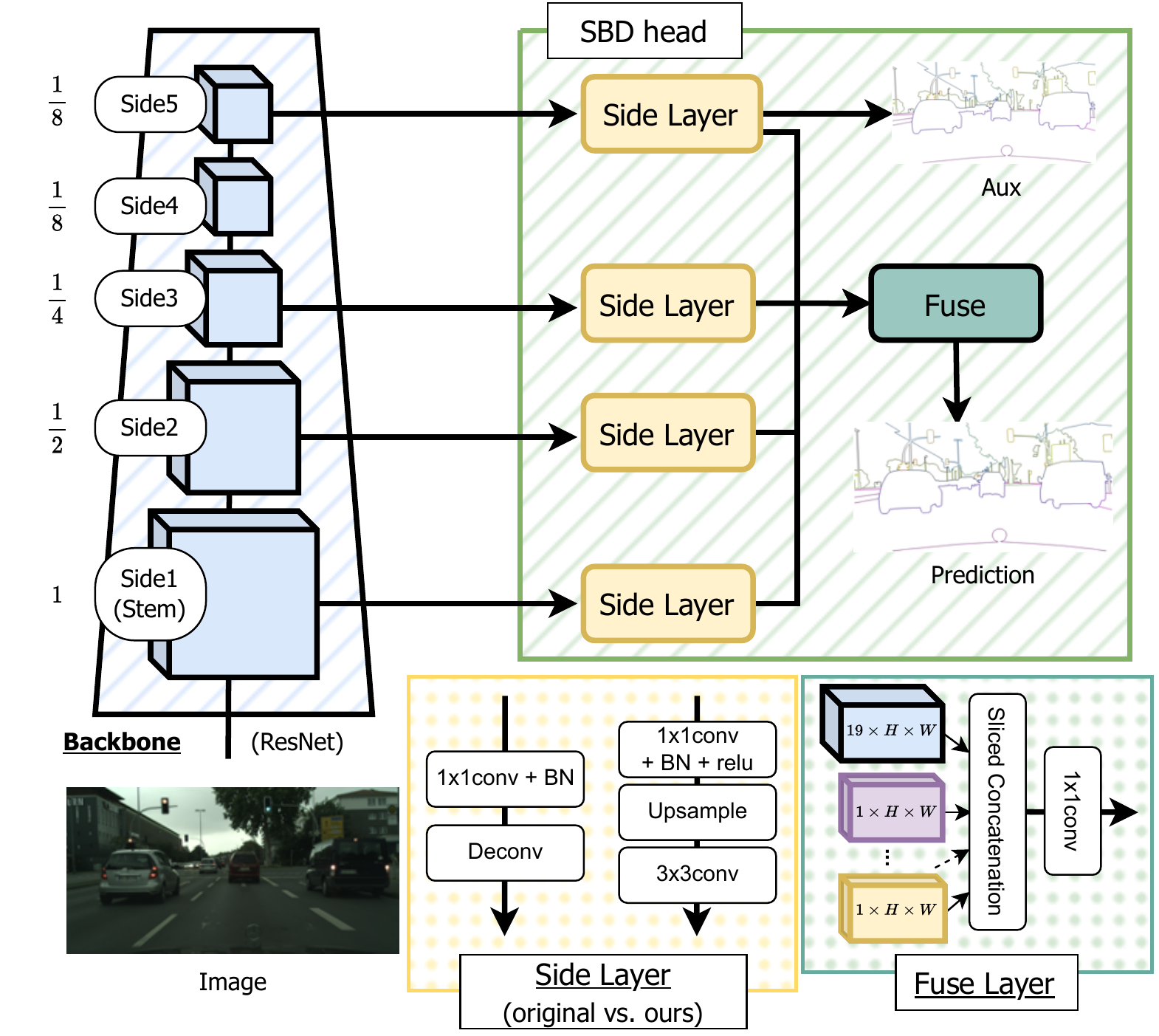}
  \caption{Overview of the CASENet Architecture. The architecture utilizes sides 1, 2, 3, and 5 of the backbone, where the Side Layer is applied. The Side Layers consist of a single convolutional layer followed by a deconvolutional layer which upsamples the feature resolution to the size of the input image. While this Side Layer works fine, the output prediction produces heavy artifacts. To mitigate the artifacts, we used a $1\times1$ convolutional kernel followed by bilinear upsampling with a $3\times3$ convolutional kernel. Finally, the features are concatenated into a single tensor using sliced concatenation and is applied to a $1\times1$ grouped convolutional kernel with the number of sides (four) as the group. We use the output of the last Side Layer as an auxiliary output which is supervised with semantic boundaries.}
  \label{fig:CASENet_diagram}
  \vspace{-1.5em}  
\end{figure}

\begin{figure}[ht]
  \centering
  \includegraphics[width=0.5\linewidth]{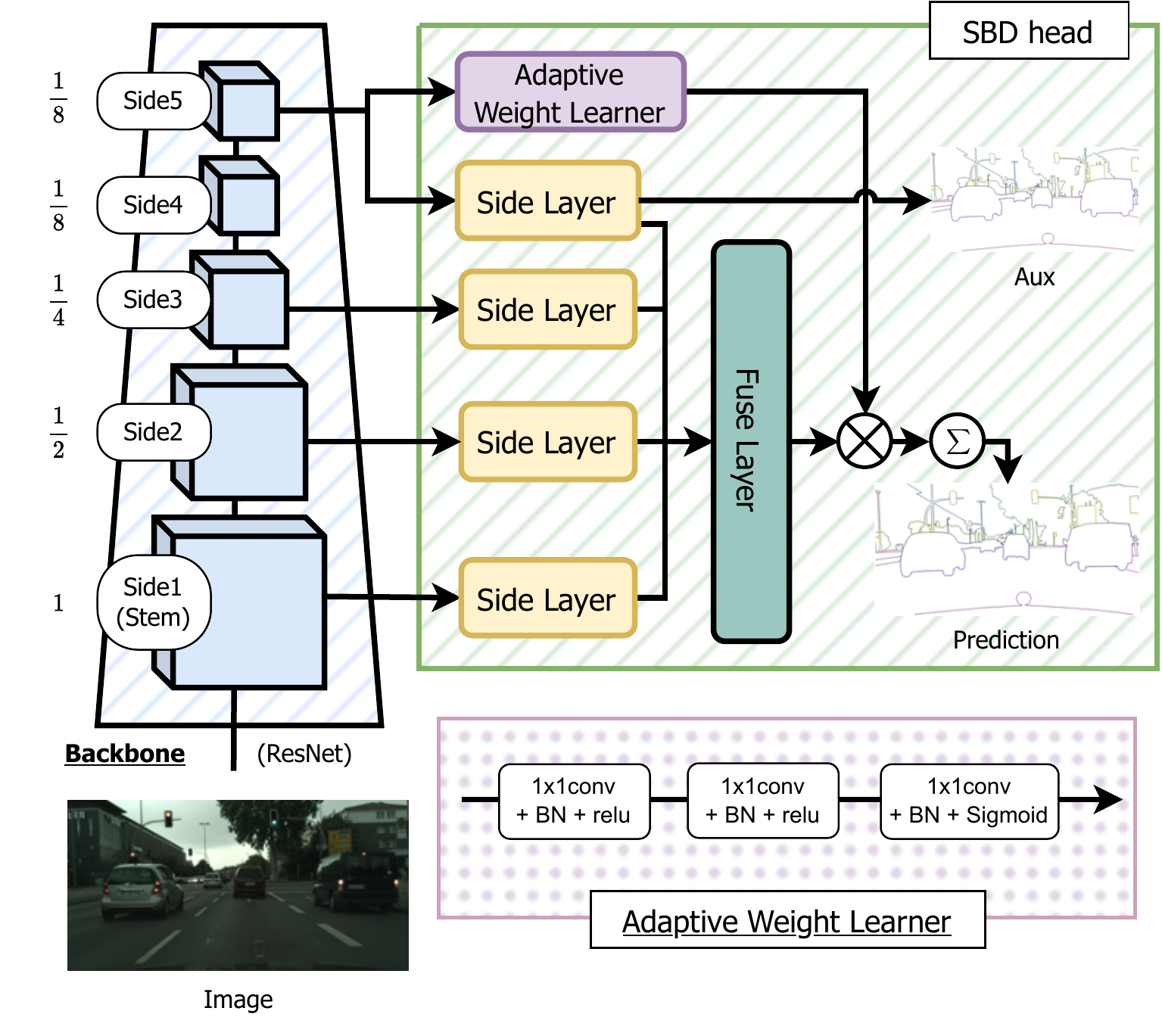}
  \caption{Overview of the DFF Architecture. The architecture adds Adaptive Weight Learner on top of the CASENet architecture that learns attentive weights, which are applied to the outputs of the Fuse Layer.}
  \label{fig:DFF_diagram}
  \vspace{-1.5em}  
\end{figure}

\begin{figure}[ht]
  \centering
  \includegraphics[width=0.5\linewidth]{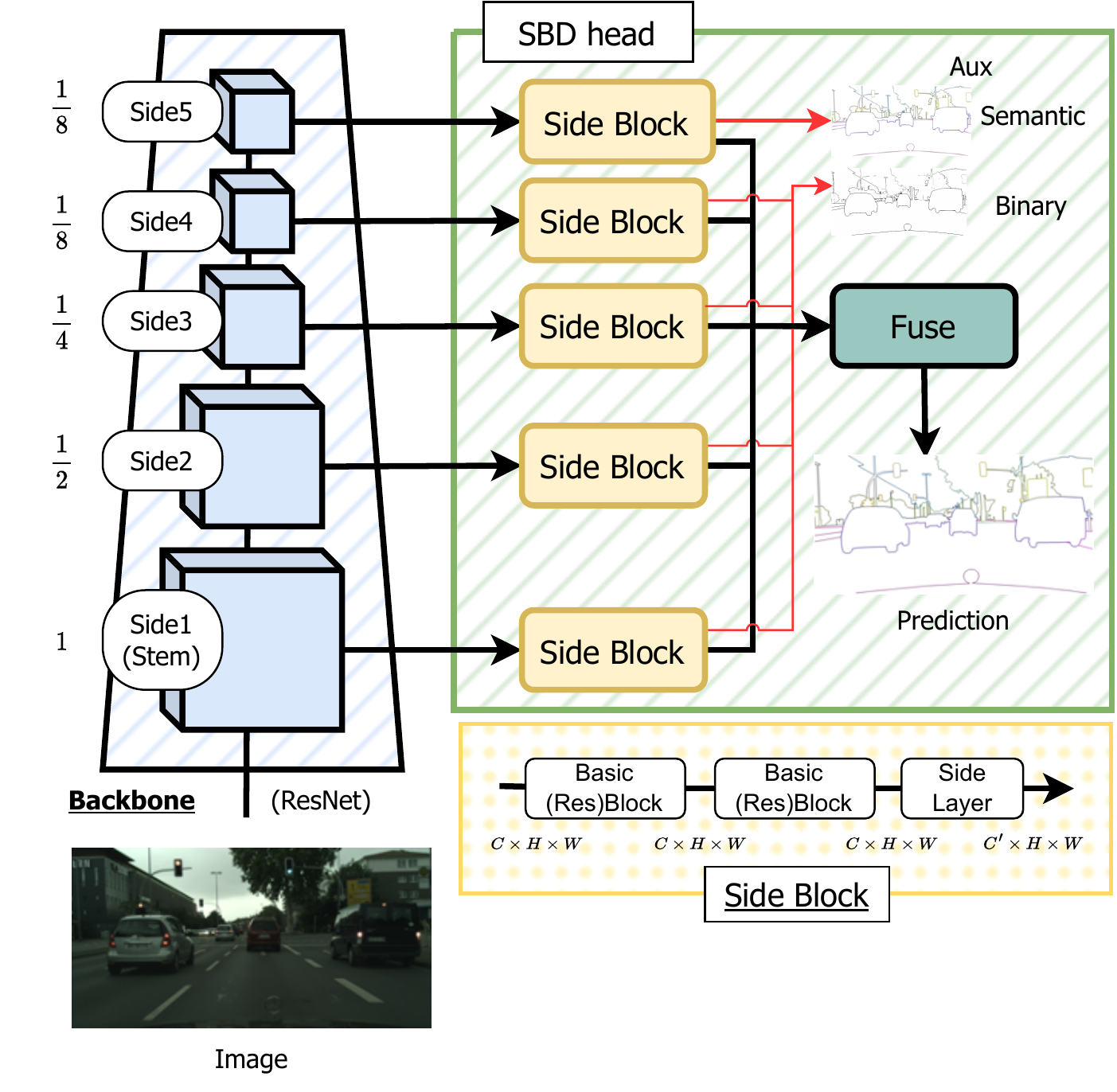}
  \caption{Overview of the DDS Architecture. The DDS architecture is similar to CASENet with additional usage of Side4 features from the backbone as well as a deeper side layer called the Side Block. The Side Block consists of two Basic ResBlocks followed by the Side Layer used in CASENet. Instead of only using the fifth Side Block as auxiliary loss, the architecture utilizes a deep supervision method where the four earlier outputs of the Side Blocks are supervised by binary boundaries as well.}
  \label{fig:DDS_diagram}
  \vspace{-1.5em}  
\end{figure}

\section{Approach}
\label{sec:approach}

The overview of the Semantic Boundary Conditioned Backbone (SBCB) framework is shown in Figure \ref{fig:intro_diagram}.
During training, we add a semantic boundary detection (SBD) head to the backbone, which receives multi-scale features from selected stages of the backbone.
The SBD head is supervised using ground-truth (GT) semantic boundaries that are generated on-the-fly using the GT segmentation masks.
During inference, if the targeted task does not require SBD, the SBD head can be discarded, resulting in a semantic segmentation model with no increase in parameters.

In Section \ref{sec:approach_sbd_heads}, we will go over existing SBD architectures and introduce the SBD heads that we will use in our experiments.
In Section \ref{sec:approach_framework}, we will go into detail about the framework by applying the SBCB framework to DeepLabV3+ and HRNet.
In Section \ref{sec:approach_otf_sbd}, we will explain the OTF semantic boundary generation module, which is the key to making this framework flexible and easy to use.
Finally, in Section \ref{sec:loss_function}, we will explain the loss function used for the framework.

\subsection{Semantic Boundary Detection Heads}
\label{sec:approach_sbd_heads}

In this section, we review some major SBD models based on ConvNets that have come out over the years.
This section will help readers understand the SBD head used in the SBCB framework as well as the experiments.
We also provide some helpful modifications that we have found worked well during our reimplementation.
Finally, we also introduce the ``Generalized'' versions of these SBD heads that we use in the SBCB framework.

\noindent\textbf{CASENet.}
The CASENet architecture was proposed by Yu et al. \cite{yu2017casenet}, which suggested a novel nested architecture without deep supervision on ResNet.
The architecture is depicted in Figure \ref{fig:CASENet_diagram}.
The ResNet backbone is modified to capture features with larger resolution (explained in depth in Section \ref{sec:albation_backbone_trick}).
At each stage of the backbone except for stage 1, the features are passed into the Side Layer, which consists of $1\times1$ convolutional kernel followed by a deconvolutional layer to increase the resolution to match the input image.
Throughout the paper, we use ``Stage'' and ``Side'' interchangeably.
Stages are based on the original papers of the backbone, oftentimes not including the Stem.
We use ``Side'', a term used in SBD-related papers, which includes Stem.
The last Side Layer (Side 5) outputs an $N_{cat}\times H \times W$ tensor while the other Side Layers (Side 1 to 4) will output $1\times H \times W$, where $N_{cat}$ is the number of categories, and $H$ and $W$ are height and width of the image.
The outputs of the Side Layers are followed by a Fuse layer which consists of a sliced concatenation of each feature with $1\times1$ convolution kernel to output an $N_{cat} \times H \times W$ a logit, which is supervised by semantic boundaries.
The output of the last Side Layer is also supervised by semantic boundaries, which are used as an auxiliary signal.
The details for semantic boundary supervision loss $\Lb_{SBD}$ for Fuse Layer and the last Side Layer is explained in Section \ref{sec:loss_function}.

We noticed that the original implementation of the Side Layer produces boundaries with heavy checkerboard artifacts and replaced the Side Layers with bilinear upsampling followed by a $3\times3$ convolutional kernel as shown in Figure \ref{fig:CASENet_diagram}.
This technique was introduced for generative models using deconvolution \cite{odena2016deconvolution}, and we modified it to not increase the number of parameters.

\noindent\textbf{DFF.}
The DFF architecture was proposed in \cite{hu2019dff} to improve the CASENet architecture by introducing the Adaptive Weight Learner to refine the output of the Fuse layer with attentive weights.
As shown in Figure \ref{fig:DFF_diagram}, the Fuse layer outputs the sliced concatenated features, and instead of a $1\times1$ convolutional kernel, the weights obtained by the Adaptive Weight Learner are applied to the tensor and summed so that the output tensor is $N_{cat} \times H \times W$.

\noindent\textbf{DDS.}
The most recent method which outperforms CASENet and DFF is called DDS, which was introduced in \cite{liu2022dds}.
DDS introduced a deeper Side Layer, known as the Side Block, which is composed of two ResNet Basic Blocks followed by a Side Layer.
The overview of the network is shown in Figure \ref{fig:DDS_diagram}.
Although CASENet avoids deep supervision of the earlier side outputs, DDS explicitly supervises all the Side Blocks.
The last output is supervised by semantic boundaries, and the earlier outputs are supervised by binary boundaries.

\begin{figure}[ht]
  \centering
  \includegraphics[width=0.5\linewidth]{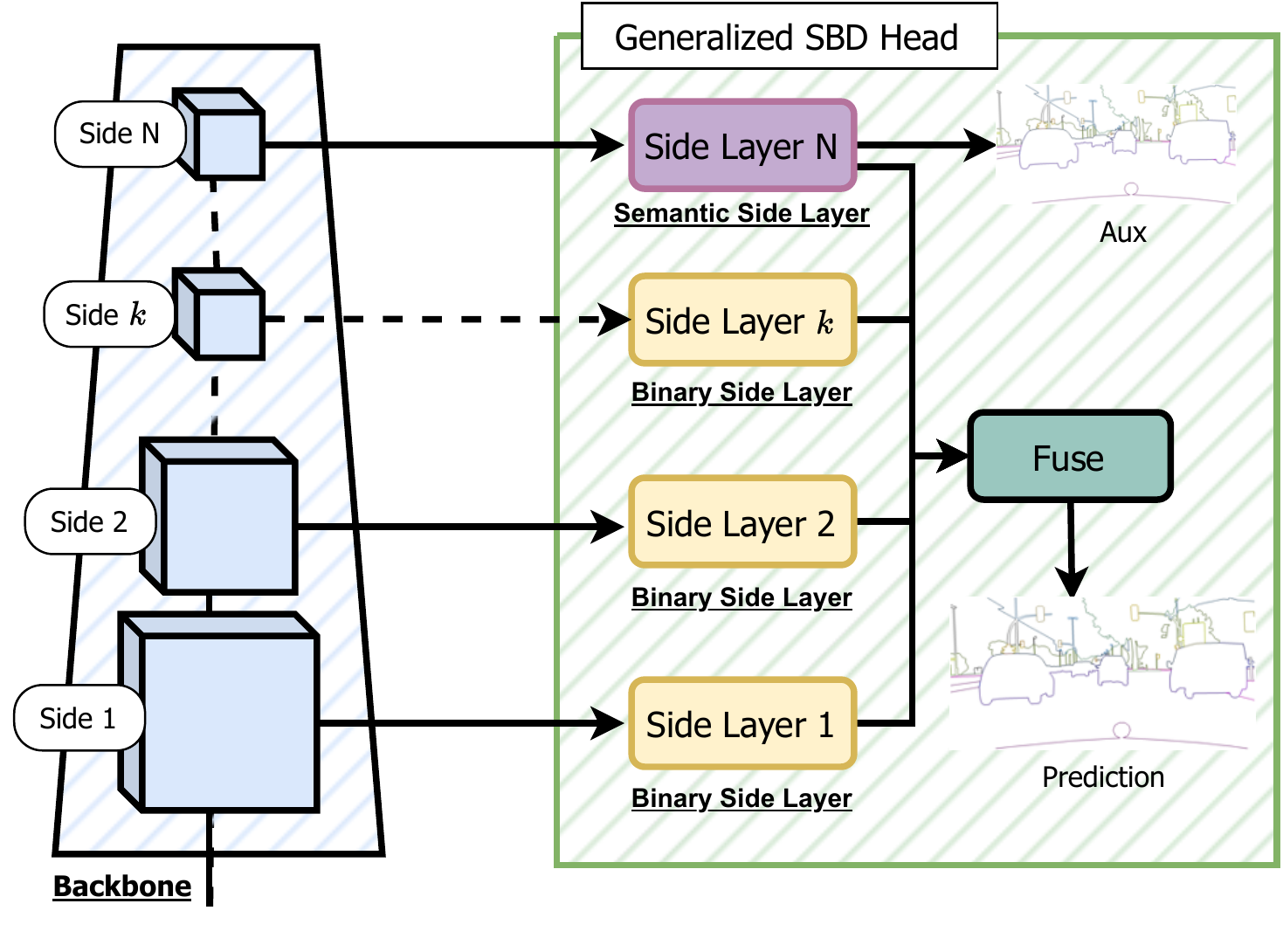}
  \caption{Overview of the Generalized CASENet Architecture. This architecture is a ``generalized'' version of the CASENet architecture in Figure \ref{fig:CASENet_diagram}. The last ($N$th) Side Layer is called the Semantic Side Layer and the input feature is called the Semantic Side. The $1\sim (N-1)$th Side Layer is called the Binary Side Layer since the input side feature (called the Binary Side) only has a single channel (like the other SBD architectures). With the generalization of not having a constrained number of sides and Side Layers, we can apply this SBD head to various backbones. This generalization can be applied to DFF and DDS architectures as well.}
  \label{fig:generalized_diagram}
  \vspace{-1.2em}  
\end{figure}

\noindent\textbf{Generalized SBD heads.} 
To facilitate the SBCB framework, we generalize the SBD heads to be applied to various backbones and segmentation architectures.
We call this SBD head the Generalized SBD head, as shown in Figure \ref{fig:generalized_diagram}.
In our framework, we generalized the architecture to have flexible Side and Fuse layers to apply any previously mentioned SBD heads (CASENet, DFF, and DDS).
The Side Layer could be the Side Layers introduced in CASENet or the Side Blocks in DDS.
The Fuse Layer could be the Fuse Layer introduced in CASENet or the Fuse Layer with Adaptive Weight Learner in DFF.
The number of Sides is also flexible where semantic boundaries supervise the $Nth$ side output with binary boundaries supervising the earlier side outputs when DDS is used.

\begin{figure}[ht]
  \centering
  \includegraphics[width=0.5\linewidth]{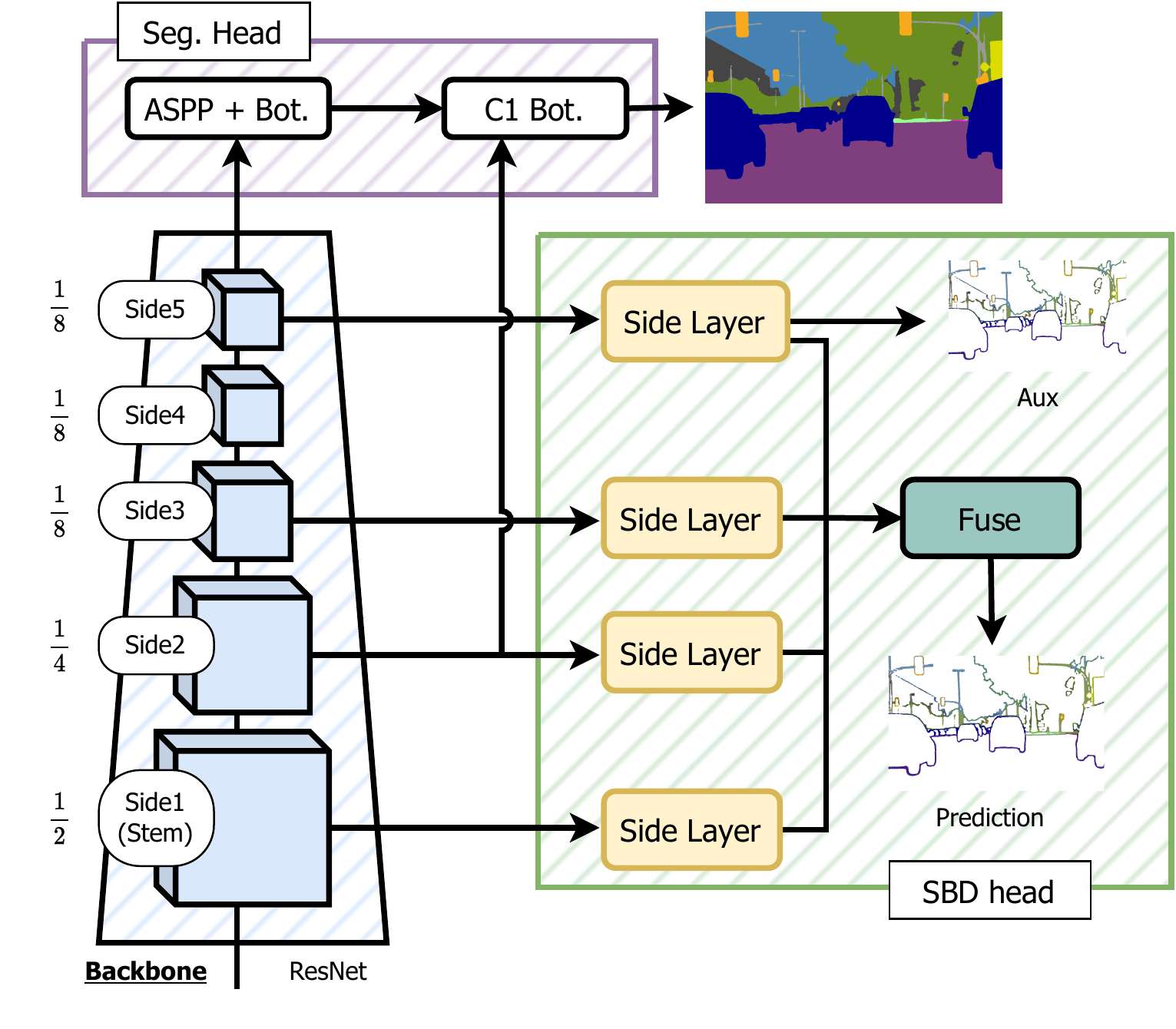}
  \caption{Diagram showcasing how the SBCB framework is applied to DeepLabV3+ segmentation head.}
  \label{fig:deeplabv3plus_sbcb_diagram}
  \vspace{-1.5em}  
\end{figure}

\begin{figure}[ht]
  \centering
  \includegraphics[width=0.5\linewidth]{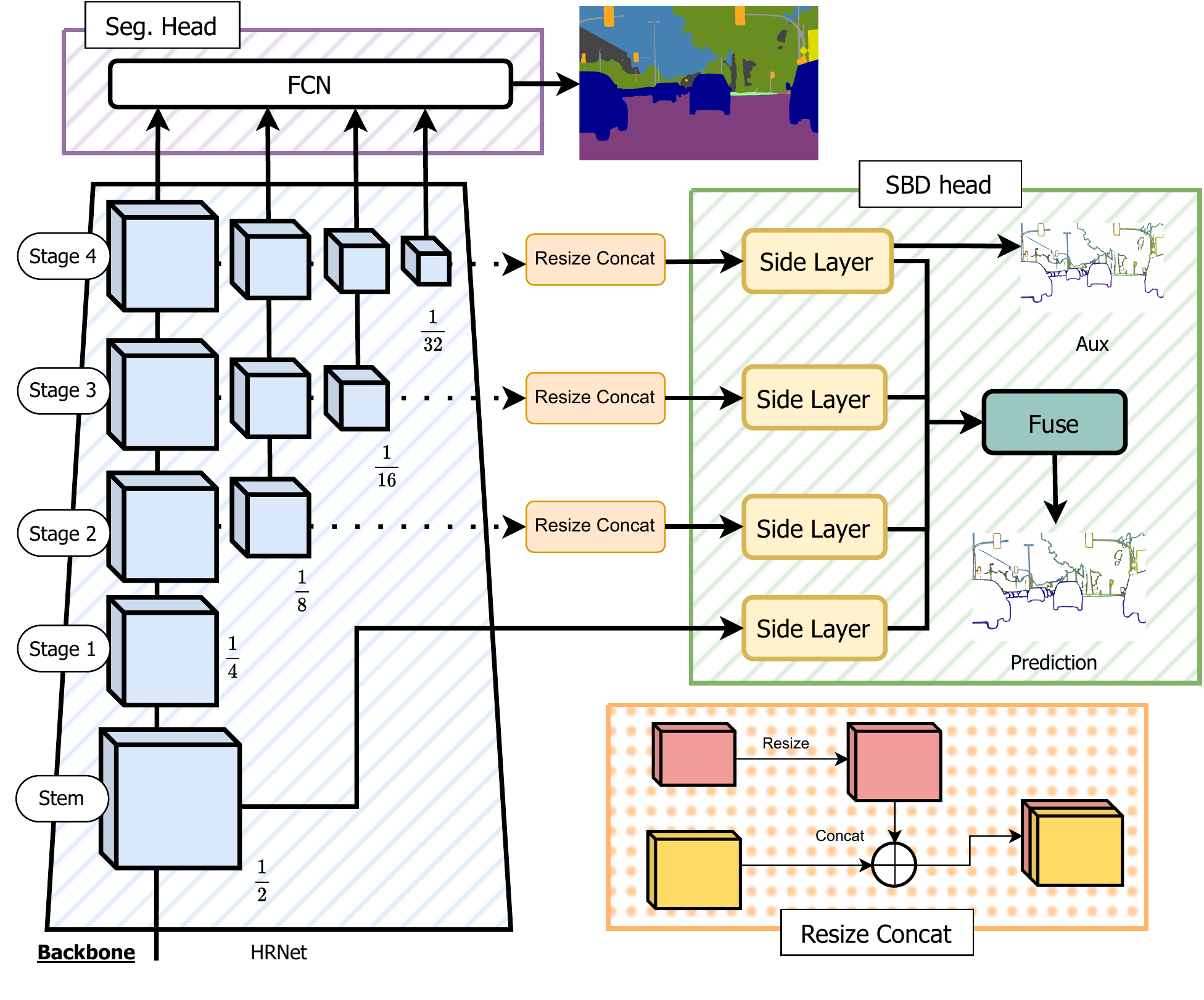}
  \caption{Diagram showcasing how the SBCB framework is applied to HRNet backbone with FCN segmentation head.}
  \label{fig:hrnet_sbcb_diagram}
  \vspace{-1.5em}  
\end{figure}

\subsection{Framework}
\label{sec:approach_framework}

In this section, we will introduce how we apply the SBD heads we reviewed in Section \ref{sec:approach_sbd_heads} for the SBCB framework.
To make the framework more comprehensive, we will provide case studies of applying the SBCB framework to popular architectures such as DeepLabV3+ and HRNet.
The SBCB framework can be applied similarly to the other architectures, and we will explore this in Section \ref{sec:applications}.

\noindent\textbf{DeepLabV3+ + SBCB.}
To apply the SBCB framework to DeepLabV3+, we do not need to adjust the number of Side Layers since the backbone is ResNet as shown in Figure \ref{fig:deeplabv3plus_sbcb_diagram}.
We take the features from each side and use them for the SBD head.
The general method of applying the SBCB framework will not change for different SBD heads.
For example, when applying the DDS head, we take the Side 4 features and change the Side Layers to Side Blocks.

\noindent\textbf{HRNet + SBCB.}
The HRNet backbone is composed of four stages, as shown in Figure \ref{fig:hrnet_sbcb_diagram}.
Since the first stage already reduces the resolution to $1/4$, we use the features from the stem for the first Side Layer.
The HRNet differs from ResNet in that the feature resolutions are consistent throughout the stages while branching out into smaller resolutions in each stage.
Because of this, we resize and concatenate the features of each stage before feeding it through the Side Layer.
We take all the features of each stage to motivate better conditioning of the backbone.

To apply the SBCB framework to different backbone architectures, we must consider the following,
\begin{itemize}
  \item Does the first Side Layer receive features with the largest resolution?
  \item Are any features not being utilized at each Side or Stage?
  \item Which Side or Stage is best suited for semantic boundary supervision?
\end{itemize}
When applying SBCB to hierarchical backbones like ResNet, the earlier stages should be applied to binary side layers, while the last layer is naturally suited for the semantic side layer.
Fortunately, most semantic segmentation architectures use some sort of hierarchical backbones, which makes applying the SBCB framework simple.
When we have backbones such as HRNet, where features are hierarchical and branching out, we must make sure to incorporate all of the features; i.e., concatenate.
For heavily customized backbones, like the ones we will explore in Section \ref{sec:applications}, we can still apply the SBCB framework by considering the three key items.
Some backbones that are developed for classification tasks may downsample the feature resolution.
It may be beneficial to increase the feature resolution by changing the strides and dilations of the convolutional kernel, so the first side feature has the resolution of at least a $1/2$ of the input image.
For this, we can apply the ``backbone trick,'' which we will discuss in Section \ref{sec:albation_backbone_trick}.

\begin{figure}[ht]
  \centering
  \includegraphics[width=0.5\linewidth]{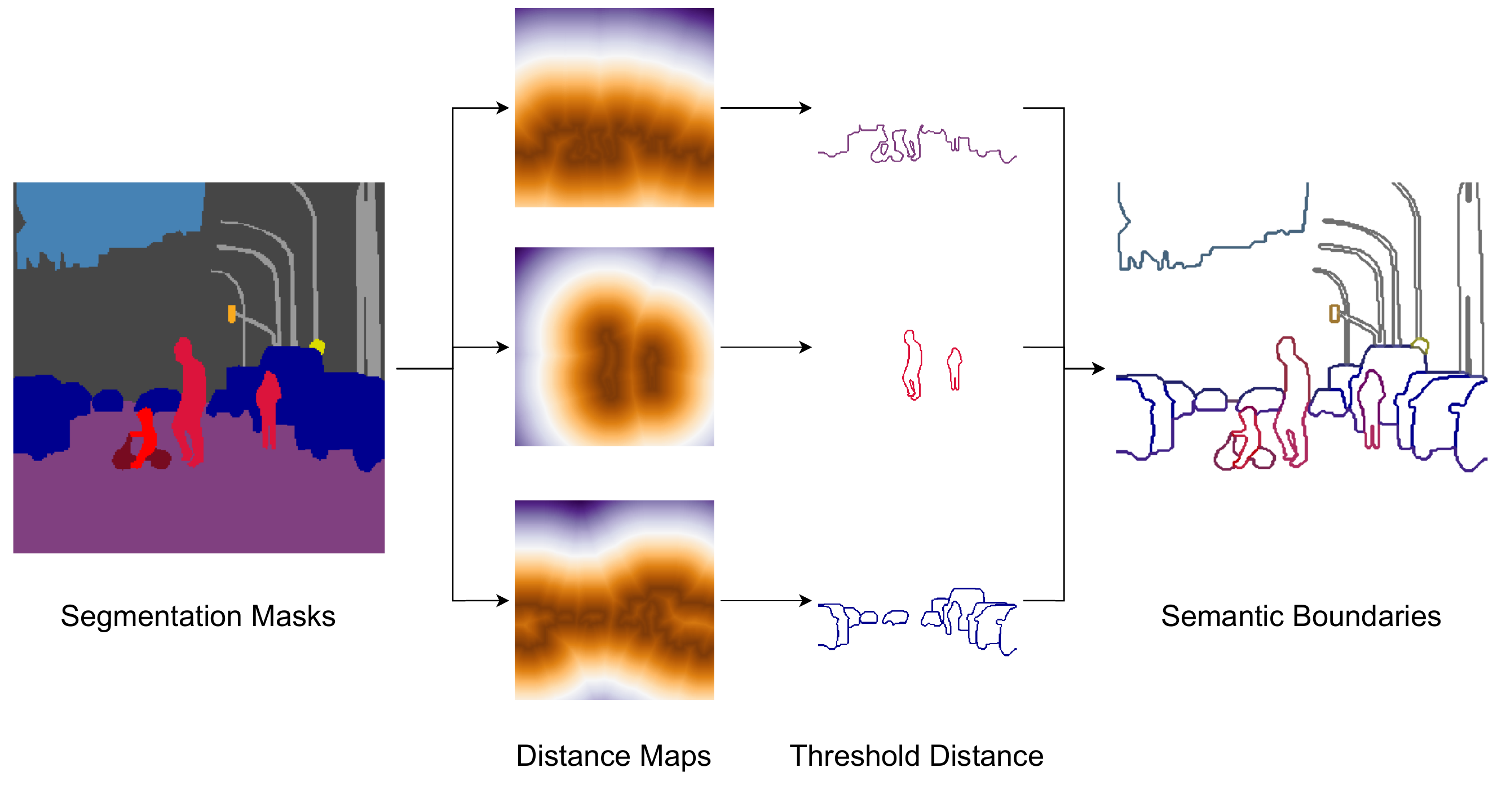}
  \caption{Overview of the OTFGT module. We apply the signed distance function to segmentation masks to obtain category-specific distance maps. We then threshold the distances by the radius of the boundaries to obtain category-specific boundaries. The boundaries are concatenated to form a semantic boundary tensor for supervision.}
  \label{fig:otfgt_diagram}
  \vspace{-1.5em}  
\end{figure}

\begin{figure}[ht]
  \begin{subfigure}[t]{0.5\linewidth}
    \includegraphics[width=\textwidth]{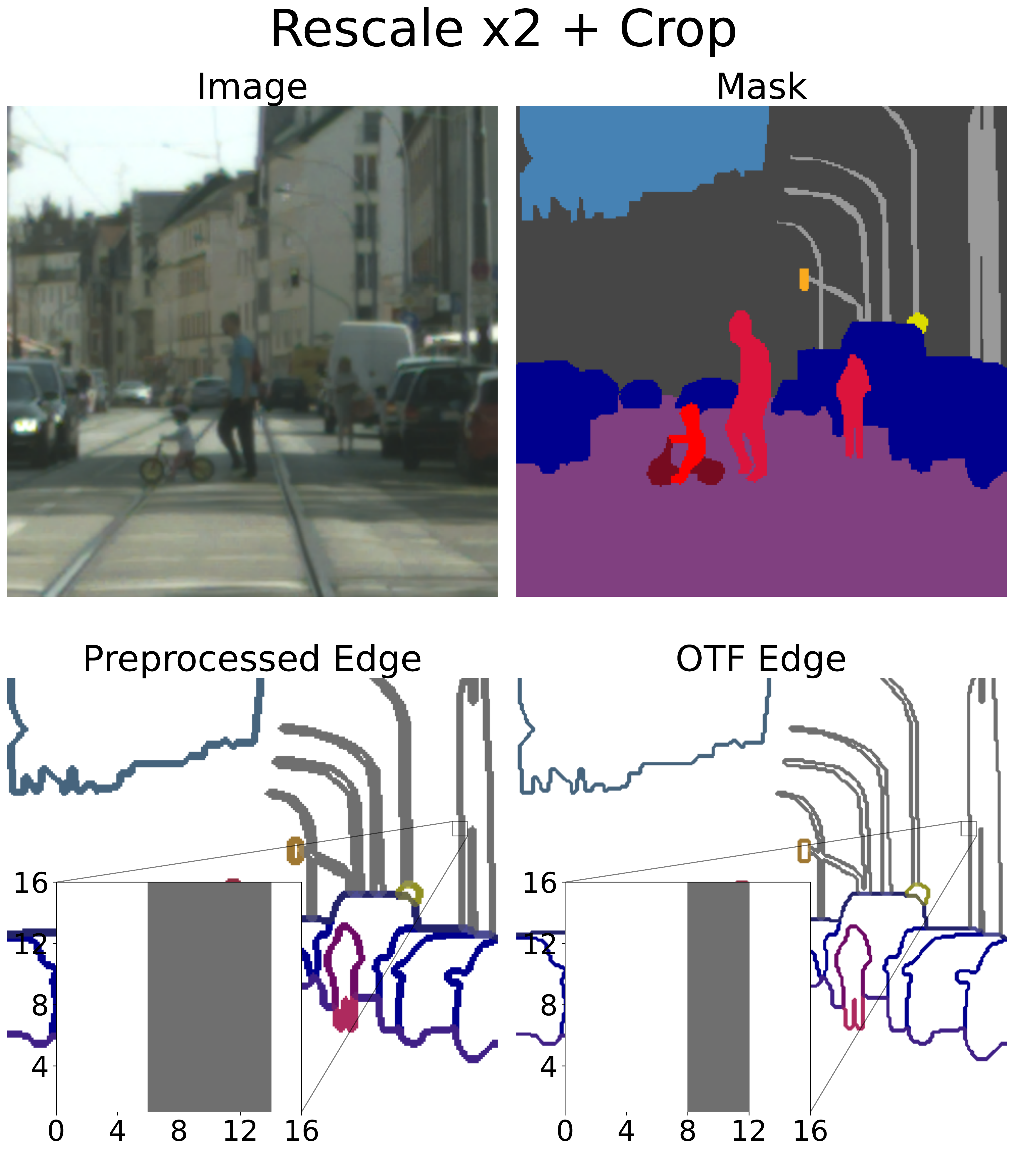}
    \caption{Data rescaled by a factor of 2.}
    \label{fig:boundary_rescale_double}
  \end{subfigure}%
  \hfill
  \begin{subfigure}[t]{.5\linewidth}
    \includegraphics[width=\textwidth]{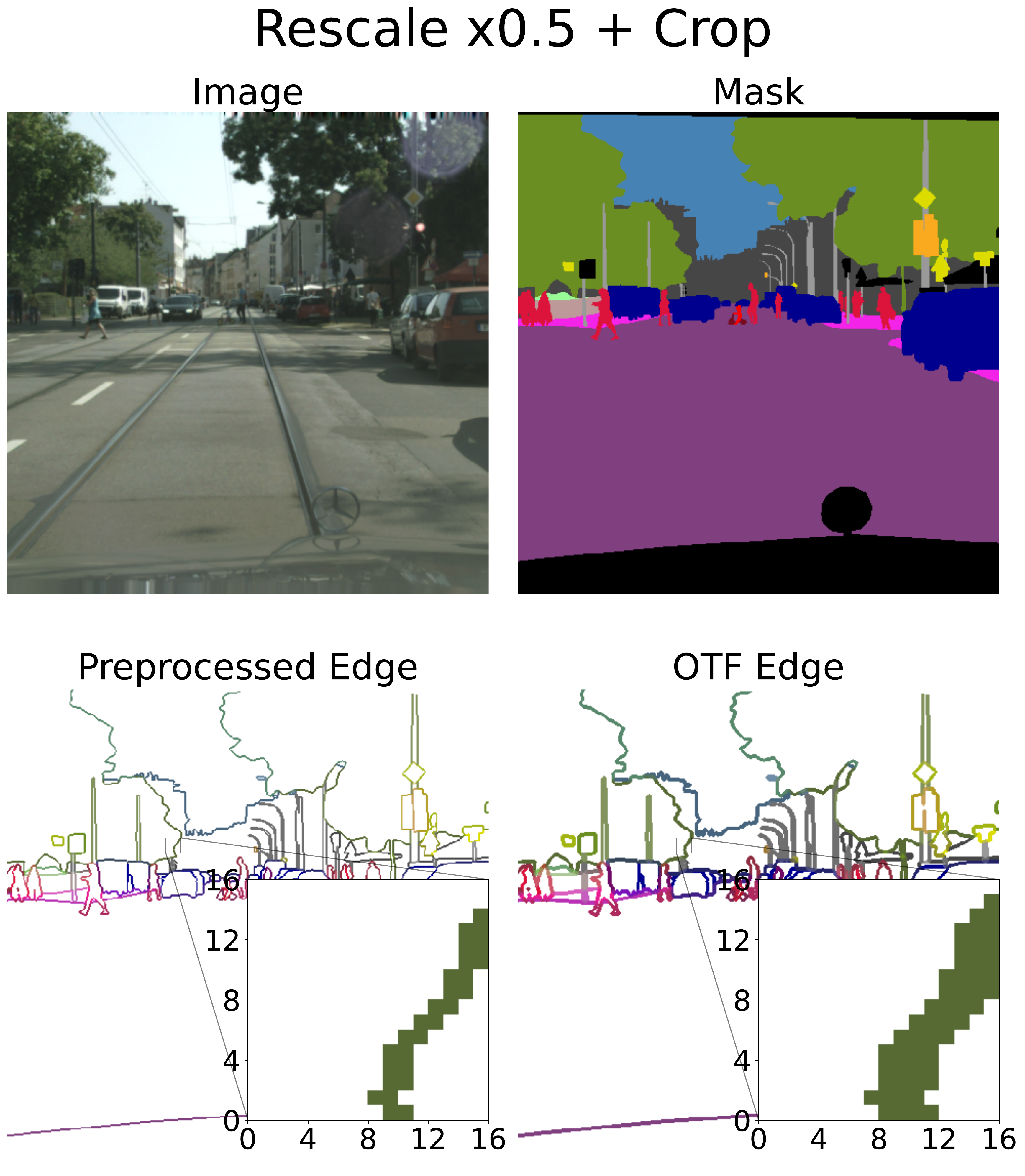}
    \caption{Data rescaled by half.}
    \label{fig:boundary_rescale_half}
  \end{subfigure}%
  \caption{The two figures represents sample validation images, masks, and boundaries from the Cityscapes validation split, which we rescale and crop to $512\times512$. In each of the figures, we compare the two methods of preprocessing. The one on the left uses preprocessed boundaries, and the one on the right uses OTFGT boundaries. We can see that OTFGT boundaries have consistent boundary widths, while preprocessed boundaries will vary depending on the rescale value.}
  \label{fig:otf_boundary_comparisons}
  \vspace{-1.5em}  
\end{figure}

\subsection{On-the-fly Ground Truth Generation}
\label{sec:approach_otf_sbd}

For the task of SBD and edge detection, humans manually annotate the edges.
Thus, the annotated image's scales and the width of the edges are predetermined.
Some datasets for SBD, such as the Cityscapes dataset and SBD dataset, provide the preprocessing of GT boundaries from semantic and instance masks to provide more training data.
Nevertheless, the number of scales is limited since it is infeasible to generate various scales before training.
On the other hand, in the semantic segmentation task, it is a common practice to resize and rescale the GT mask during training to remedy overfitting by increasing the variations of the dataset.
This is impossible for semantic boundaries since resizing will result in inconsistent edge widths, as shown in Figure \ref{fig:otf_boundary_comparisons}.

To remedy this, we developed a simple semantic boundary generation algorithm that is efficient enough to run in the preprocessing pipeline called the on-the-fly (OTF) semantic boundary GT generation module (OTFGT).
The OTFGT generates semantic boundaries from semantic segmentation masks and can create instance-sensitive boundaries when instance segmentation masks are available.
The details of the OTFGT are explained in Appendix \ref{sec:appendix_generation}.

\subsection{Loss Functions.}
\label{sec:loss_function}

Given an input image, the model generates segmentation and boundary maps with pre-defined semantic categories.
We apply cross-entropy (CE) loss, $\Lb_{Seg}$, for each pixel of the segmentation map.
As for the SBD head, we apply binary cross entropy (BCE) loss for multi-label boundaries, $\Lb_{SBD}$ following \cite{yu2017casenet}.
While CASENet and DFF use only multi-label boundaries for supervision, DDS also introduces deep supervision of edges where earlier side outputs are supervised with binary boundary maps using BCE loss $\Lb_{Bdry}$ \cite{liu2022dds}.
Generally, the loss function used is,
\begin{equation}
\Lb = \Lb_{Seg} + \alpha\sum^{S_{SBD}}\Lb_{SBD} + \beta\sum^{S_{Bin}}\Lb_{Bdry},
\label{eq:loss}
\end{equation}
where $\alpha$ and $\beta$ are constants for balancing the effects losses from each task.
$S_{SBD}$ is a set of semantic boundary predictions and $S_{Bin}$ is a set of binary boundary predictions.
For CASENet and DFF, $S_{SBD} = \{\bf{B}_{side N}, \bf{B}_{fuse}\}$, where $\bf{B}_{side N}$ represents the last side output and $\bf{B}_{fuse}$ represents the final fused prediction as shown in Figure \ref{fig:generalized_diagram}.
For DDS, we supervise $S_{SBD} = \{\bf{B}_{side N}, \bf{B}_{fuse}\}$ and $S_{Bin} = \{\bf{B}_{side k}, \ldots \bf{B}_{side 2}, \bf{B}_{side 1}\}$.


\begin{figure*}[t]
  \centering
  \includegraphics[width=0.7\linewidth]{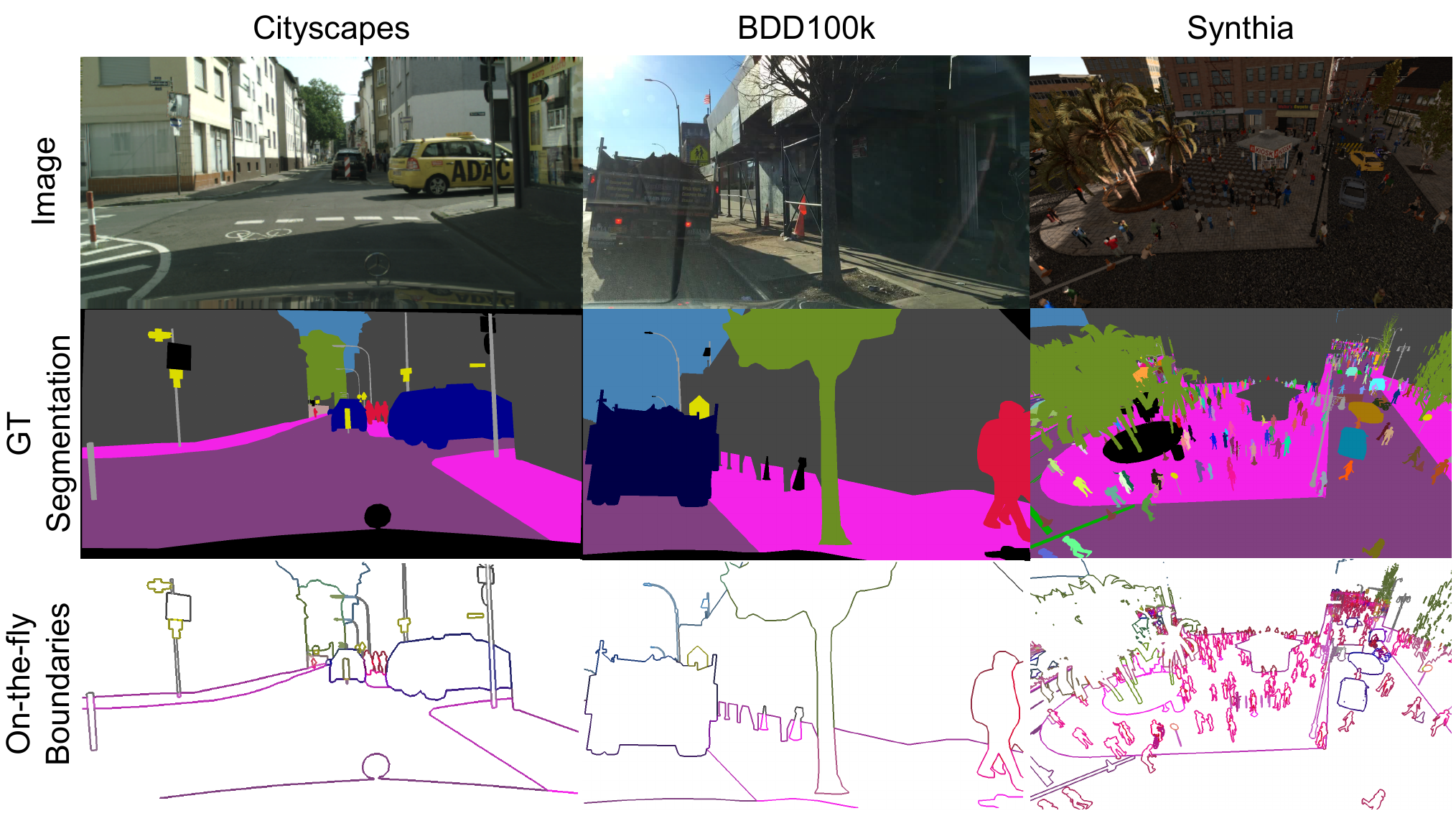}
  \caption{The three main datasets that we used for the experiments. We show a sample input image, segmentation GT, and the result of OTF semantic boundary generation for each dataset. Humans annotate the Cityscapes and BDD100K datasets, and the segmentation masks are clean but tend to have imperfections around the boundaries and exhibit ``polygon'' masks. On the other hand, the Synthia dataset is data from a game engine, and the annotations are pixel-perfect, making this a challenging dataset for semantic segmentation. The segmentation mask for Synthia also contains instance segmentation, which is used in the OTF semantic boundary generation but not for the segmentation task. The BDD100K and Synthia datasets are less widely used than the Cityscapes dataset. However, the BDD100K and Synthia datasets contain more variations of natural noise and corruptions (weather, heavy light reflections, etc...) which will help benchmark the methods fairly. The images are best seen in color and zoomed in.}
  \label{fig:datasets}
  \vspace{-1.5em}  
\end{figure*}

\section{Experiment Setup}
\label{sec:setup}

In this section, we go over the details of our experiments, including the dataset, hyperparameters, and implementations.

\subsection{Datasets}
\label{sec:setup_datasets}
In our experiments, we use three datasets, namely Cityscapes, BDD100K, and Synthia datasets.
We visualize and explain each dataset in Figure \ref{fig:datasets}.

\noindent\textbf{Cityscapes.}
We evaluate our models on the popular Cityscapes dataset \cite{cordts2016cityscapes}, which contains 2975 training images, 500 validation images, and 1525 testing images with $19$ semantic categories.
Following \cite{yu2017casenet,yu2018seal,hu2019dff,liu2022dds}, the dataset has also been widely adopted as the standard benchmark for SBD.
We conduct quantitative studies for both semantic segmentation and SBD on the validation set and benchmark our method on the test set for semantic segmentation.

\noindent\textbf{BDD100K.}
The BDD100K dataset \cite{Yu2018BDD100K} is a driving dataset that is aimed at multi-task learning for autonomous driving.
This dataset is the largest driving video dataset with 100K video frames and ten tasks, and it contains 10K images with a resolution of $1280\times720$ for the semantic segmentation task.
The dataset is split into 7K training, 1K validation, and 2K test splits, for which we only use the training and validation split for our ablation experiments.
The annotated labels are the same as the Cityscapes dataset.

\noindent\textbf{Synthia.}
The Synthia dataset \cite{Ros2016Synthia} is a CG dataset generated using a simulator aimed at providing auxiliary datasets for Cityscapes as well as for experimenting with domain adaptation.
We use the ``Rand'' set of the dataset, which contains 13.4K images with a resolution of $1280\times760$ with annotated categories that are the same as the Cityscapes dataset.
We use Synthi as a stand-alone dataset to explore the effect of the SBCB framework under annotations with precise boundaries.
We split the dataset into 10.4K training, 1.5K validation, and 1.5K test split.

\subsection{Evaluation Metrics}
\label{sec:setup_metrics}

\noindent\textbf{Segmentation Metrics.}
We consider the mean of intersection-over-union (mIoU) for evaluating the segmentation performances.
Following \cite{takikawa2019gscnn}, we adopt boundary F-score to evaluate the segmentation performance around the boundary of the masks.
We use a pixel width of 3px for boundary F-score unless explicitly stated.

\noindent\textbf{Boundary Detection Metrics.}
We follow \cite{yu2018seal} and adopt the maximum F-score (mF) at the optimal dataset scale (ODS) evaluated on the instance-sensitive "thin" protocol for SBD.

\subsection{Implementation Details}
\label{sec:setup_impl_details}

\noindent\textbf{Data Loading.}
Unless explicitly stated, we unify the experiments' training crop size, training iterations, and batch size for both tasks to $512\times1024$, 40k, and 8, respectively, for the Cityscapes dataset.
We used the same parameters for Synthia and BDD100K datasets but used a crop size of $640\times640$.
We fine-tuned the models evaluated in the Cityscapes test benchmark for an additional 40k iterations using the training and validation split, following the works of \cite{yu2021coupled}.
We perform common data augmentations, notably random scaling (scale factors in $[0.5, 2.0]$), horizontal flip, and photo-metric distortions.

\noindent\textbf{Optimization.}
We employ the SGD optimizer with a momentum coefficient of $0.9$ and a weight decay coefficient of $5\times10^{-4}$ during training.
We optimize the network by using the "poly" learning rate policy where the initial learning rate ($0.01$) is multiplied by $(1 - \frac{iter}{max\_iter})^{power}$ with $power=9$.

\noindent\textbf{Loss.}
We set $\alpha=5$ and $\beta=1$ for our loss function in Eq. \ref{eq:loss}.

\noindent\textbf{Inference.}
In our experiments, we conduct evaluations with single-scale whole inference for the Cityscapes dataset and slide inference for Synthia and BDD100K datasets.
For evaluating semantic segmentation performance in Section \ref{sec:applications_cityscapes}, we apply multi-scale and flip (MS+Flip) inference strategy with scales of $[0.5, 0.75, 1.0, 1.25, 1.5, 1.75, 2.0]$.

\noindent\textbf{Software and Hardware.}
To conduct all of our experiments, we use PyTorch and modify the popular semantic segmentation framework ``mmsegmentation'' \cite{mmseg2020} for our task.
We reported all experimental results using the same software and hardware and trained all models under the same conditions.
The models are trained on two NVIDIA A6000 GPUs and evaluated on a single NVIDIA RTX8000.


\begin{table*}[t]
  \caption{Ablation studies to compare SBD heads as auxiliary signals. The number of parameters and performance for SBCB is the model used for training. For inference, the number of parameters and performance equals the baseline. Unless explicitly stated, the hyperparameters for training are the same throughout the experiments (crop size of $512\times1024$ and ResNet-101 backbone.)}
  \label{table:ablation_which_head}
  \begin{subtable}[t]{0.48\textwidth}
    \scriptsize
    \caption{Results on the Cityscapes validation split.}
    \setlength\tabcolsep{4pt}
    \centering
    \begin{tabular}[t]{c||cc|cc}
    Head & mIoU & mF (ODS) & Param. & GFLOPs \\
    \hline\hline
    DeepLabV3+          &  79.5 & - & 60.2M   & 506  \\
    \hline
    CASENet             & \multirow{3}{*}{-} & 63.7 & 42.5M  & 357   \\
    DFF                 & & 65.5 & 42.8M  & 395    \\
    DDS                 & & 73.4 & 243.3M & 2079  \\
    \hline
    SBCB (CASENet)      & 80.3 & 74.4 & 60.2M  & 508 \\
    SBCB (DFF)          & 80.2 & 74.6 & 60.5M  & 545 \\
    SBCB (DDS)          & \textbf{80.6} & \textbf{75.8} & 261.0M & 2228 \\
    \end{tabular}
    \label{table:ablation_sbd_heads}
  \end{subtable}
  \hspace{\fill}
  \begin{subtable}[t]{0.48\textwidth}
    \flushright
    \caption{Results on the Cityscapes validation split. The crop size is set to $769\times769$.}
    \scriptsize
    \setlength\tabcolsep{4pt}
    \centering
    \begin{tabular}[t]{c||cc|cc}
    Head & mIoU & mF (ODS) & Param. & GFLOPs \\
    \hline\hline
    DeepLabV3+          & 78.9 & - & 60.2M   & 506  \\
    \hline
    CASENet             & \multirow{3}{*}{-} & 68.6 & 42.5M  & 357   \\
    DFF                 & & 68.9 & 42.8M  & 395    \\
    DDS                 & & 75.5 & 243.3M & 2079  \\
    \hline
    SBCB (CASENet)      & 80.3  & 74.0 & 60.2M  & 508 \\
    SBCB (DFF)          & 80.0  & 74.8 & 60.5M  & 545 \\
    SBCB (DDS)          & \textbf{80.4}  & \textbf{75.6} & 261.0M & 2228 \\
    \end{tabular}
    \label{table:ablation_crop_size_769}
  \end{subtable}

  \bigskip

  \begin{subtable}[t]{0.40\textwidth}
    \caption{Results using HRNet-48 (HR48) backbone on the Cityscapes validation split.}
    \scriptsize
    \setlength\tabcolsep{4pt}
    \centering
    \begin{tabular}{c||cc|cc}
    Head & mIoU & mF (ODS) & Param. & GFLOPs \\
    \hline\hline
    FCN                 & 80.5 & - & 65.9M & 187 \\
    \hline
    CASENet             & \multirow{3}{*}{-} & 75.7 & 65.3M  & 172 \\
    DFF                 & & 75.3 & 65.5M  & 210 \\
    DDS                 & & 78.9 & 89.0M & 946 \\
    \hline
    SBCB (CASENet)      & \textbf{82.0} & 78.9 & 65.9M & 187 \\
    SBCB (DFF)          & 81.5 & 78.8 & 66.0M & 221 \\
    SBCB (DDS)          & 81.0 & \textbf{79.3} & 89.5M & 1012 \\
    \end{tabular}
    \label{table:ablation_backbone_hr48}
  \end{subtable}
  \vspace{\fill}
  \begin{subtable}[t]{0.28\textwidth}
    \flushright
    \caption{Results on the BDD100K validation split.}
    \scriptsize
    \setlength\tabcolsep{4pt}
    \centering
    \begin{tabular}{c||cc}
      Head & mIoU & mF (ODS) \\
    \hline\hline
      DeepLabV3+     & 60.0 & -  \\
    \hline
      CASENet        & \multirow{3}{*}{-} & 55.7 \\
      DFF            &                    & 57.3 \\
      DDS            &                    & 59.9 \\
    \hline
      SBCB (CASENet) & 61.4 & 56.6 \\
      SBCB (DFF)     & 62.0 & 58.1 \\
      SBCB (DDS)     & \textbf{64.1} & \textbf{60.2} \\
    \end{tabular}
    \label{table:ablation_resnet101_bdd100k}
  \end{subtable}
  \vspace{\fill}
  \begin{subtable}[t]{0.28\textwidth}
    \flushright
    \caption{Results on the Synthia dataset.}
    \scriptsize
    \setlength\tabcolsep{4pt}
    \centering
    \begin{tabular}{c||cc}
      Head & mIoU & mF (ODS) \\
    \hline\hline
      DeepLabV3+     & 74.5 & -  \\
    \hline
      CASENet        & \multirow{3}{*}{-} & 61.0 \\
      DFF            &                    & 64.8 \\
      DDS            &                    & \textbf{67.6} \\
    \hline
      SBCB (CASENet) & \textbf{75.9} & 65.2 \\
      SBCB (DFF)     & 75.3 & 66.5 \\
      SBCB (DDS)     & 75.7 & 67.0 \\
    \end{tabular}
    \label{table:ablation_resnet101_synthia}
  \end{subtable}
  \vspace{-1.2em}  
\end{table*}

\begin{figure*}[t]
  \centering
  \includegraphics[width=0.95\linewidth]{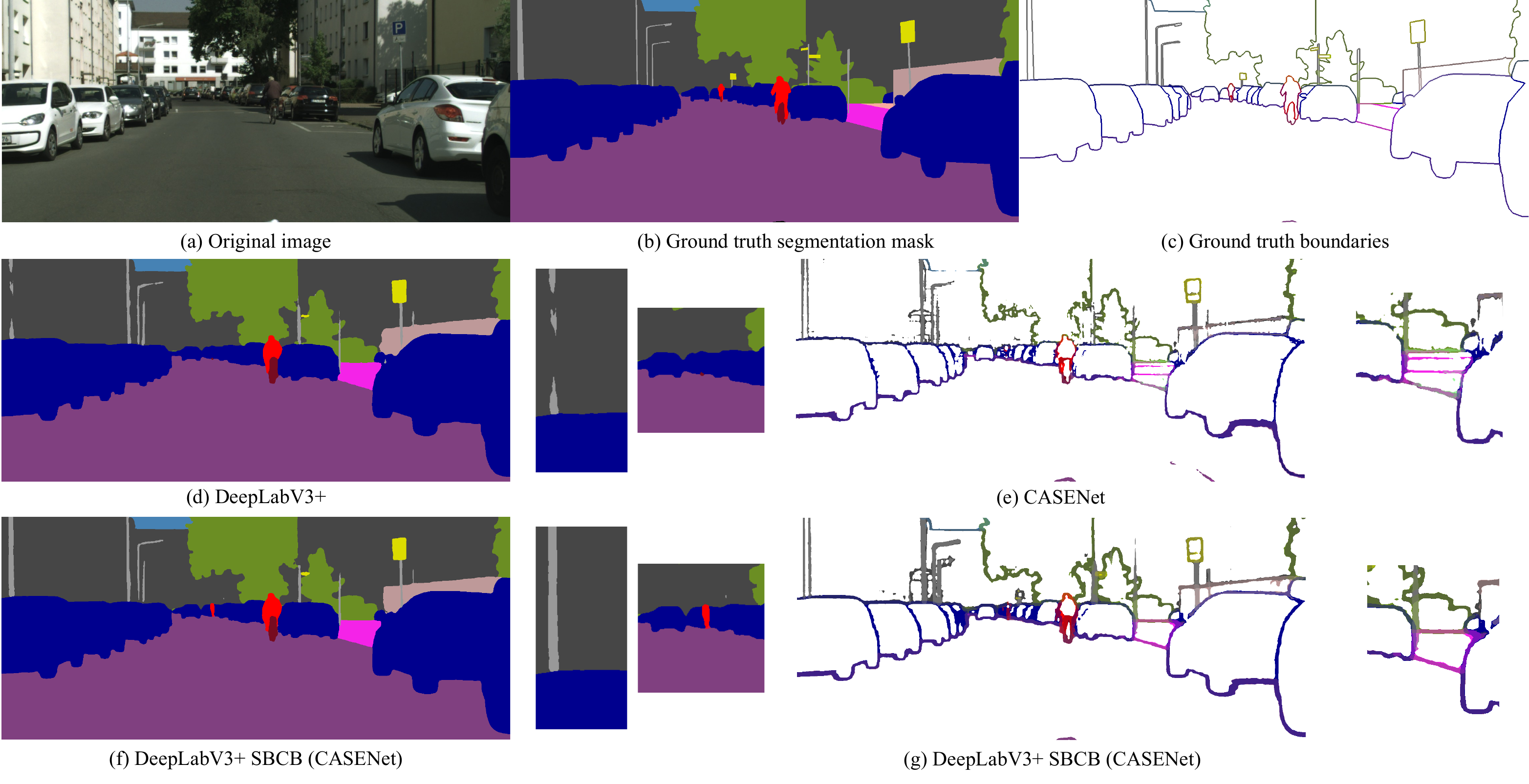}
  \caption{Overview of the task along with predictions of baseline methods and the model trained with the SBCB framework (CASENet). (a), (b), and (c) are the input image, ground-truth (GT) segmentation map, and GT semantic boundary map. Note that because the task of SBD is pixel-wise multi-label classification, the visualized semantic boundary maps have overlapped boundaries. In (d), we show the output of DeepLabV3+, a popular semantic segmentation model. The semantic boundary detection (SBD) baseline is CASENet, which we show in (e). The output of DeepLabV3+ trained with the SBCB framework using the CASENet head, which we show in (f) and (g). We can see that small and thin objects are recognized better using the framework and smoother boundaries with fewer artifacts.}
  \label{fig:ablation_qualitative}
  \vspace{-1.2em}  
\end{figure*}

\begin{figure*}[t]
  \centering
  \includegraphics[width=0.9\linewidth]{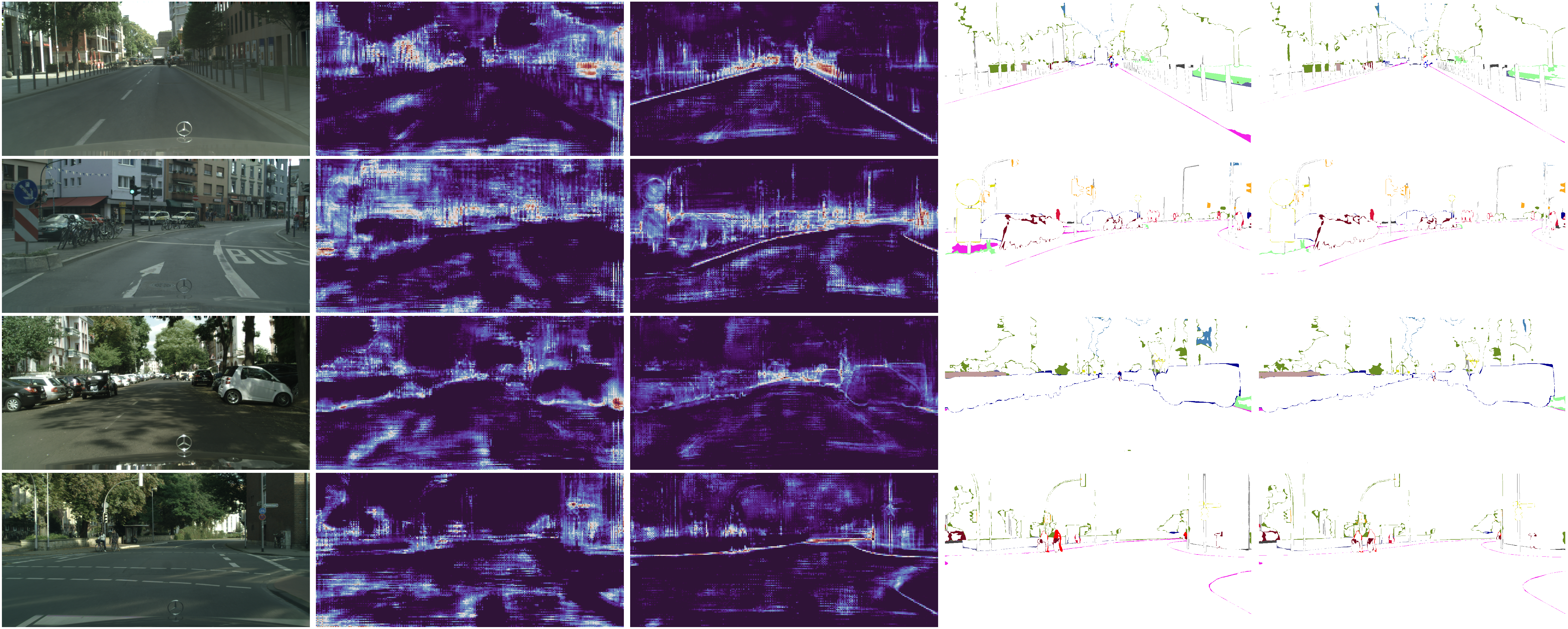}
  \caption{Visualization of the backbone features and segmentation errors of DeepLabV3+ with and without the SBCB framework. From the left, each column represents the input image, last-stage features without SBCB, last-stage features with SBCB, segmentation errors without SBCB, and segmentation errors with SBCB. As we can see, the features learned using the SBCB framework exhibits boundary-aware characteristic because it is conditioned on semantic boundaries. Consequently, this results in better segmentation, especially around the mask boundaries. Best seen in color and zoomed in.}
  \label{fig:ablation_deeplabv3plus_vis}
  \vspace{-1.5em}  
\end{figure*}

\section{Ablation Studies}
\label{sec:ablations}

In this section, we perform ablation studies using the SBCB framework in various aspects.
In Section \ref{sec:ablation_head}, we compare the SBD heads and choose a candidate for experimenting throughout the paper.
In Section \ref{sec:ablation_sides}, we figure out the optimal side configuration.
In Section \ref{sec:ablation_per_category}, we look at which categories benefit the most from the SBCB framework.
In Section \ref{sec:ablation_auxiliary_signals}, we compare the SBCB framework with other auxiliary tasks.
In Sections \ref{sec:ablation_gscnn} and \ref{sec:ablation_segfix}, we compare the SBCB framework with the state-of-the-art multi-task and post-processing method and show that our framework can complement the methods to further improving the segmentation quality.
In Section \ref{sec:albation_backbone_trick}, we investigate the effects of modifying the backbone configuration in a simple yet effective way to improve segmentation and SBD.
In Section \ref{sec:ablation_sbd}, we show the effects of the SBCB framework on the task of SBD.
Finally, in Section \ref{sec:ablation_fscore}, we show that our framework improves segmentation around the boundaries.

\begin{table*}[ht]
\caption{Per-category IoU for the Cityscapes validation split.}
\scriptsize
\setlength\tabcolsep{2.5pt}
\centering
\begin{tabular}{cc|c|ccc ccc ccc ccc ccc ccc c}
Method & SBCB & mIoU & road & swalk & build. & wall & fence & pole & tlight & sign & veg & terrain & sky & person & rider & car & truck & bus & train & motor & bike \\
\hline\hline
\multirow{3}{*}{PSPNet} &            & 77.6 & 98.0 & 83.9 & 92.4 & 49.5 & 59.3 & 64.5 & 71.7 & 79.0 & 92.4 & 64.2 & 94.7 & 81.8 & 60.5 & 95.0 & 77.8 & 89.1 & 80.1 & 63.4 & 77.9 \\
                        & \checkmark & 78.7 & 98.3 & 85.7 & 92.7 & 52.7 & 60.7 & 66.3 & 72.7 & 80.8 & 92.8 & 64.3 & 94.6 & 82.4 & 62.7 & 95.3 & 79.5 & 88.6 & 81.4 & 66.0 & 78.7 \\
\cdashline{3-22}\rule{0pt}{2.5ex} 
                        &            & \textcolor{red}{+1.1} & \textcolor{red}{+0.3} & \textcolor{red}{+1.8} & \textcolor{red}{+0.3} & \textcolor{red}{+3.2} & \textcolor{red}{+1.4} & \textcolor{red}{+1.8} & \textcolor{red}{+1.0} & \textcolor{red}{+1.8} & \textcolor{red}{+0.4} & \textcolor{red}{+0.1} & \textcolor{blue}{-0.1} & \textcolor{red}{+0.6} & \textcolor{red}{+2.2} & \textcolor{red}{+0.3} & \textcolor{red}{+1.7} & \textcolor{blue}{-0.5} & \textcolor{red}{+1.3} & \textcolor{red}{+2.6} & \textcolor{red}{+0.8} \\
\hline
\multirow{3}{*}{DeepLabV3} &            & 79.2 & 98.1 & 84.6 & 92.6 & 54.5 & 61.7 & 64.6 & 71.7 & 79.3 & 92.6 & 64.6 & 94.6 & 82.4 & 63.8 & 95.4 & 83.2 & 90.9 & 84.2 & 67.7 & 78.1 \\
                           & \checkmark & 79.9 & 98.4 & 86.4 & 93.0 & 55.3 & 63.7 & 66.8 & 72.9 & 80.4 & 94.9 & 65.4 & 94.9 & 83.3 & 65.9 & 95.5 & 81.9 & 92.3 & 81.3 & 68.2 & 78.9 \\
\cdashline{3-22}\rule{0pt}{2.5ex} 
                           &            & \textcolor{red}{+0.7} & \textcolor{red}{+0.3} & \textcolor{red}{+1.8} & \textcolor{red}{+0.4} & \textcolor{red}{+0.8} & \textcolor{red}{+2.0} & \textcolor{red}{+2.2} & \textcolor{red}{+1.2} & \textcolor{red}{+1.1} & \textcolor{red}{+2.3} & \textcolor{red}{+0.8} & \textcolor{red}{+0.3} & \textcolor{red}{+0.9} & \textcolor{red}{+2.1} & \textcolor{red}{+0.1} & \textcolor{blue}{-1.3} & \textcolor{red}{+1.4} & \textcolor{blue}{-2.9} & \textcolor{red}{+0.5} & \textcolor{red}{+0.8} \\
\hline
\multirow{3}{*}{DeepLabV3+} &            & 79.5 & 98.1 & 85.0 & 92.9 & 53.2 & 62.8 & 66.5 & 72.1 & 80.4 & 92.7 & 64.9 & 94.7 & 82.8 & 63.6 & 95.5 & 85.1 & 90.9 & 82.2 & 69.4 & 78.4 \\
                            & \checkmark & 80.3 & 98.3 & 85.9 & 93.4 & 65.7 & 65.6 & 68.5 & 73.0 & 81.4 & 92.8 & 66.1 & 95.3 & 83.3 & 65.6 & 95.5 & 81.3 & 88.3 & 78.1 & 68.7 & 78.8 \\
\cdashline{3-22}\rule{0pt}{2.5ex} 
                            &            & \textcolor{red}{+0.8} & \textcolor{red}{+0.2} & \textcolor{red}{+0.9} & \textcolor{red}{+0.5} & \textcolor{red}{+12.5} & \textcolor{red}{+2.8} & \textcolor{red}{+2.0} & \textcolor{red}{+0.9} & \textcolor{red}{+1.0} & \textcolor{red}{+0.1} & \textcolor{red}{+1.2} & \textcolor{red}{+0.6} & \textcolor{red}{+0.5} & \textcolor{red}{+2.0} & 0 & \textcolor{blue}{-3.8} & \textcolor{blue}{-2.6} & \textcolor{blue}{-4.1} & \textcolor{blue}{-0.7} & \textcolor{red}{+0.4} \\
\end{tabular}
\label{table:ablation_per_category}
\vspace{-1.2em}  
\end{table*}

\begin{table}[hb]
  \caption{Results using ResNet-101 backbone with different sides on Cityscapes validation split.}
  \scriptsize
  \setlength\tabcolsep{4pt}
  \centering
  \begin{tabular}{c|l||cc}
    Head & \multicolumn{1}{c||}{Sides} & mIoU & $\Delta$ \\
  \hline\hline
    \multirow{5}{*}{PSPNet}     &                   & 77.6 & \\
    \cdashline{2-4}\rule{0pt}{2.5ex} 
                                & 1 + 5             & 78.5 & +0.9 \\
                                & 1 + 2 + 5         & 78.6 & +1.0 \\
                                & 1 + 2 + 3 + 5     & 78.7 & \textbf{+1.1} \\
                                & 1 + 2 + 3 + 4 + 5 & 78.5 & +0.9 \\
    \hline
    \multirow{5}{*}{DeepLabV3}  &                   & 79.2 & \\
    \cdashline{2-4}\rule{0pt}{2.5ex} 
                                & 1 + 5             & 79.8 & +0.6 \\
                                & 1 + 2 + 5         & 79.9 & \textbf{+0.7} \\
                                & 1 + 2 + 3 + 5     & 79.9 & \textbf{+0.7} \\
                                & 1 + 2 + 3 + 4 + 5 & 79.4 & +0.2 \\
    \hline
    \multirow{5}{*}{DeepLabV3+} &                   & 79.5 & \\
    \cdashline{2-4}\rule{0pt}{2.5ex} 
                                & 1 + 5             & 80.1 & +0.6 \\
                                & 1 + 2 + 5         & 80.1 & +0.6 \\
                                & 1 + 2 + 3 + 5     & 80.3 & +0.8 \\
                                & 1 + 2 + 3 + 4 + 5 & 80.5 & \textbf{+1.0} \\
  \end{tabular}
  \label{table:ablation_sides_cityscapes}
  \vspace{-1.2em}  
\end{table}

\subsection{Which SBCB head to use?}
\label{sec:ablation_head}

In this section, we explore the effects of using different semantic boundary detection (SBD) heads for the SBCB framework and find the best candidate for further evaluation.

Table \ref{table:ablation_sbd_heads} shows the DeepLabV3+ model trained using three different SBD heads, CASENet, DFF, and DDS, compared with single-task baseline models.
All SBD heads for the SBCB framework improve the single-task DeepLabV3+ model.
We also can see that the joint training helps improve the SBD metric (maximum F-score).
We also included the number of parameters and computational costs in GFLOPs to show how much the SBD heads can introduce costs during training.
While DDS adds high costs for training, it is also the most performant of the three heads.
On the other hand, CASENet only adds a few number of parameters to the original model.
The trade-off of using DDS over CASENet for the SBCB framework might not be beneficial in terms of performance gains, which will be more evident as we evaluate DDS on other datasets and backbones.

In Figure \ref{fig:ablation_qualitative}, we show qualitative results of the CASENet head applied to DeeplabV3+ compared with the baselines.
We can see that the additional semantic boundary supervision allows the model to detect smaller thin objects better.
We can also see that the SBCB framework allows for better boundary detection with fewer artifacts and better perception of objects.

\noindent\textbf{Different crop size.}
In semantic segmentation, crop size is one of the most important hyperparameter, and we test the SBD heads on $769\times769$, another popular crop size.
The results are shown in Table \ref{table:ablation_crop_size_769}, where the general trend is the same as the results from Table \ref{table:ablation_sbd_heads}.

\noindent\textbf{Different backbone.}
We also explore the effects of using another popular backbone, namely HRNet-48 (HR48), and the results are shown in Table \ref{table:ablation_backbone_hr48}.
This time, we can see that the CASENet head outperforms DDS and DFF by significant margins ($1.0\%$ and $0.5\%$, respectively).
The CASENet head also achieves mF of $78.9\%$, identical to the heavy and inefficient single-task DDS model.

\noindent\textbf{Different datasets.}
In computer vision, the model's performance differs depending on the dataset.
We additionally evaluate the SBD heads on the BDD100K dataset and Synthia, as shown in Tables \ref{table:ablation_resnet101_bdd100k} and \ref{table:ablation_resnet101_synthia} respectively.
On the BDD100K, the DDS head significantly outperforms the baseline model and CASENet head.
The DFF head performs better than the CASENet head for this dataset for the first time.
As for Synthia, the CASENet head performs better than DDS.

\noindent\textbf{CASENet as the candidate.}
While the DDS head performs better than CASENet for the most part, when we consider the additional parameters and computational costs, it is beneficial to use the CASENet head.
Besides, the SBD head in the SBCB framework is only used as an auxiliary signal, and the CASENet head outperforms DDS in some results.
It can be noted that when it is dire to squeeze out higher metrics and the computational costs can be ignored, using the DDS head may result in better metrics.
For the rest of the paper, we use the CASENet head as our main SBD head for the SBCB framework.

In Figure \ref{fig:ablation_deeplabv3plus_vis}, we show qualitative visualizations that compare DeepLabV3+ with and without the CASENet head.
We can see from the feature maps obtained from the last stage of the backbone that the backbone conditioned on SBD exhibits boundary-aware characteristics, which reduces the segmentation errors, especially around the boundaries.

\begin{table*}[t]
  \caption{Ablation studies to compare different backbone conditioning methods. We investigate the effects on three popular segmentation heads: PSPNet, DeepLabV3, and DeepLabV3+. Note that all of the methods use ResNet-101 as the backbone. The number of parameters and performance for SBCB is the model used for training. For inference, the number of parameters and performance equals the baseline.}
  \begin{subtable}[t]{0.48\textwidth}
    \caption{Results on the Cityscapes validation split.}
    \scriptsize
    \setlength\tabcolsep{4pt}
    \centering
    \begin{tabular}{c|ccc|c||cc}
      Head & FCN & BBCB & SBCB & Param. & mIoU & $\Delta$ \\
    \hline\hline
      \multirow{6}{*}{PSPNet}     &            &            &            & 65.58M &  77.6 & \\
      \cdashline{2-7}\rule{0pt}{2.5ex} 
                                  & \checkmark &            &            & +2.37M &  78.3 & +0.7 \\
                                  &            & \checkmark &            & +0.01M &  78.1 & +0.5 \\
                                  &            &            & \checkmark & +0.05M &  \textbf{78.7} & +1.1 \\
      \cdashline{2-7}\rule{0pt}{2.5ex} 
                                  & \checkmark & \checkmark &            & +2.37M &  79.1 & +1.5 \\
                                  & \checkmark &            & \checkmark & +2.41M &  \textbf{79.4} & +1.8 \\
      \hline
      \multirow{6}{*}{DeepLabV3}  &            &            &            & 84.72M &  79.2 & \\
      \cdashline{2-7}\rule{0pt}{2.5ex}
                                  & \checkmark &            &            & +2.37M &  79.3 & +0.1 \\
                                  &            & \checkmark &            & +0.01M &  79.6 & +0.4 \\
                                  &            &            & \checkmark & +0.05M &  \textbf{79.9} & +0.7 \\
      \cdashline{2-7}\rule{0pt}{2.5ex} 
                                  & \checkmark & \checkmark &            & +2.37M &  \textbf{80.1} & +0.9 \\
                                  & \checkmark &            & \checkmark & +2.41M &  \textbf{80.1} & +0.9 \\
      \hline
      \multirow{6}{*}{DeepLabV3+} &            &            &            & 60.2M  &  79.5 & \\
      \cdashline{2-7}\rule{0pt}{2.5ex}
                                  & \checkmark &            &            & +2.37M &  79.7 & +0.2 \\
                                  &            & \checkmark &            & +0.01M &  79.9 & +0.4 \\
                                  &            &            & \checkmark & +0.05M &  \textbf{80.3} & +0.8 \\
      \cdashline{2-7}\rule{0pt}{2.5ex} 
                                  & \checkmark & \checkmark &            & +2.37M &  \textbf{80.6} & +1.1 \\
                                  & \checkmark &            & \checkmark & +2.41M &  80.5 & +1.0 \\
    \end{tabular}
    \label{table:ablation_aux_cityscapes}
  \end{subtable}
  \vspace{\fill}
  \begin{subtable}[t]{0.48\textwidth}
    \flushright
    \caption{Results on the Synthia dataset.}
    \scriptsize
    \setlength\tabcolsep{4pt}
    \centering
    \begin{tabular}{c|ccc||cc}
      Head & FCN & BBCB & SBCB & mIoU & $\Delta$ \\
    \hline\hline
      \multirow{6}{*}{PSPNet}     &            &            &            & 70.5 & \\
      \cdashline{2-6}\rule{0pt}{2.5ex}
                                  & \checkmark &            &            & 70.1 & -0.4 \\
                                  &            & \checkmark &            & 70.7 & +0.2 \\
                                  &            &            & \checkmark & \textbf{71.7} & +1.2 \\
      \cdashline{2-6}\rule{0pt}{2.5ex} 
                                  & \checkmark & \checkmark &            & 70.7 & +0.2 \\
                                  & \checkmark &            & \checkmark & \textbf{71.6} & +1.1 \\
      \hline
      \multirow{6}{*}{DeepLabV3}  &            &            &            & 70.9 & \\
      \cdashline{2-6}\rule{0pt}{2.5ex}
                                  & \checkmark &            &            & 70.6 & -0.3 \\
                                  &            & \checkmark &            & 70.7 & -0.2 \\
                                  &            &            & \checkmark & \textbf{71.9} & +1.0 \\
      \cdashline{2-6}\rule{0pt}{2.5ex} 
                                  & \checkmark & \checkmark &            & 70.5 & -0.4 \\
                                  & \checkmark &            & \checkmark & \textbf{71.0} & +0.1 \\
      \hline
      \multirow{6}{*}{DeepLabV3+} &            &            &            & 72.4 & \\
      \cdashline{2-6}\rule{0pt}{2.5ex}
                                  & \checkmark &            &            & 72.0 & -0.4 \\
                                  &            & \checkmark &            & 72.1 & -0.3 \\
                                  &            &            & \checkmark & \textbf{73.5} & +1.1 \\
      \cdashline{2-6}\rule{0pt}{2.5ex} 
                                  & \checkmark & \checkmark &            & 72.3 & -0.1 \\
                                  & \checkmark &            & \checkmark & \textbf{73.5} & +1.1 \\
    \end{tabular}
    \label{table:ablation_aux_synthia}
  \end{subtable}
  \vspace{-0.8em}  
\end{table*}

\subsection{Which sides to supervise?}
\label{sec:ablation_sides}

The CASENet head applied to the ResNet backbone has five sides, Sides 1, 2, 3, 4, and 5.
In Table \ref{table:ablation_sides_cityscapes}, we show the effect of using different side configurations.
For consistency with performant single-task SBD models, we constrain Side 1 and 5 because Side 1 is required for low-level understanding and has the largest feature resolution, where Side 5 is required for high-level understanding.
We added Sides 2, 3, and 4 and compared the performance gains.
Note that Sides 1+2+3+5 is the original configuration.
The table shows the original configuration works best on two models (PSPNet and DeepLabV3).
On DeepLabV3+, configuration 1+2+3+4+5 outperforms the original configuration by $0.2\%$.
We believe that the difference in performance gains is negligible, but users of the SBCB framework should know that each model could have an optimal side configuration.
Therefore, for fairness, we choose the original configuration to evaluate other models and benchmark our methods for further evaluation.

\subsection{Does it improve all categories?}
\label{sec:ablation_per_category}

Table \ref{table:ablation_per_category} provides the per-category IoU comparisons for each model.
We can see from the table that although most of the categories improve with the SBCB framework, some categories results in worse IoU.
The most frequent categories are ``truck'', ``bus'', and ``train'', which have relatively low samples and are easily confused with ``car''.
During training, additional measures, such as Online Hard Example Mining (OHEM), could mitigate this effect.

\subsection{Comparisons of different auxiliary signals}
\label{sec:ablation_auxiliary_signals}

Introduced in PSPNet \cite{zhao2017pspnet}, the authors added another classifier to the backbone to stabilize the training and improve segmentation metrics.
In detail, the authors added the FCN head to the fourth stage (one before the last stage) in the backbone.
The auxiliary FCN head is trained on the same segmentation task as the main head.
This technique is still used today and abundantly in open-source projects such as \lstinline{mmseg}.

Although not used often, various papers applied binary edge and boundary detection as an auxiliary task for semantic segmentation.
Even though the task of binary boundary detection is different from semantic segmentation, the authors found that the learned features in the edge detection head can be fused into the segmentation head.

In this section, we compare the SBCB framework with the mentioned auxiliary techniques, which we call ``FCN'' and ``Binary Boundary Conditioned Backbone (BBCB)''.
Note that BBCB is the SBCB framework but is applied to binary boundary detection instead.
We applied FCN, BBCB, and SBCB on three popular segmentation heads (PSPNet, DeepLabV3, and DeepLabV3+) and used ResNet-101 as the backbone.
The results for the Cityscapes validation split are shown in Table \ref{table:ablation_aux_cityscapes}.
While all auxiliary signals improve IoU, the models trained using the SBCB framework are consistently the best.
The improvements of SBCB compared with BBCB are around twice, proving that the task of SBD is crucial.
FCN applied on PSPNet has the most gains of $0.7\%$, but FCN has minimal impact on the other models.
The BBCB and SBCB framework can complement FCN, and the results show it can achieve higher IoU.
Another important aspect is the additional parameters these auxiliary signals bring during training.
While SBCB and BBCB only add thousands of parameters, FCN adds $2.37M$ parameters.
Considering the performance gains and the additional parameters, it is clear that boundary-based auxiliary signals provide more benefits than FCN.

We also evaluate the same models and auxiliary heads on the Synthia dataset as shown in Table \ref{table:ablation_aux_synthia}.
Surprisingly, FCN and BBCB do not add much performance gains and even have worse metrics than the baselines.
However, SBCB improves upon the baseline by over $1\%$.
It is plausible that the features learned using FCN could have conflicted with the main heads.
Compared with Cityscapes, Synthia contains precise segmentation masks rendered from a CG engine instead of human annotation.
In Synthia, classes such as ``human'' and ``bike'' will have small and thin segmentation masks, which makes this dataset difficult.
Although features learned on FCN complemented the features of the main head in Cityscapes, it appears that the FCN learned to derive a conflicted segmentation map.
It is possible because there are more layers (parameters) in the FCN head compared to SBCB or BBCB.
Ostensibly, BBCB would perform well because of its shallow (far fewer parameters than FCN) architecture, but the results are contrary.
This is because the BBCB focuses on low-level features without explicitly modeling high-level semantics.
We believe the polarity of the task resulted in the main head not receiving good features for semantic segmentation for Synthia.

The SBCB framework conditions the backbone with SBD, a challenging task focusing on low-level and requires high-level features.
The SBCB framework improves the segmentation metrics better than using FCN or binary boundaries as auxiliary signals because of the hierarchical modeling of the SBD task.

\begin{table}[ht]
  \caption{Results obtained from the Cityscapes validation split. We compared the use of SegFix with auxiliary heads (SBCB and FCN) on three popular baseline models.}
  \scriptsize
  \setlength\tabcolsep{4pt}
  \centering
  \begin{tabular}{c|l||cc}
    \multicolumn{2}{c||}{Model} & mIoU & $\Delta$ \\
  \hline\hline
    \multirow{6}{*}{PSPNet} &                       & 77.6 & \\
    \cdashline{2-4}\rule{0pt}{2.5ex} 
                            & + SegFix              & 78.8 & +1.2 \\
                            & + SBCB                & 78.7 & +1.1 \\
                            & + SBCB + FCN          & 79.4 & +1.8 \\
                            & + SBCB + SegFix       & 79.7 & +2.1 \\
                            & + SBCB + FCN + SegFix & 80.3 & +2.8 \\
  \hline
    \multirow{6}{*}{DeepLabV3} &                       & 79.2 & \\
    \cdashline{2-4}\rule{0pt}{2.5ex} 
                               & + SegFix              & 80.3 & +1.1 \\
                               & + SBCB                & 79.9 & +0.7 \\
                               & + SBCB + FCN          & 80.1 & +0.9 \\
                               & + SBCB + SegFix       & 80.8 & +1.6 \\
                               & + SBCB + FCN + SegFix & 81.0 & +1.8 \\
  \hline
    \multirow{6}{*}{DeepLabV3+} &                       & 79.5 & \\
    \cdashline{2-4}\rule{0pt}{2.5ex} 
                                & + SegFix              & 80.4 & +0.9 \\
                                & + SBCB                & 80.3 & +0.8 \\
                                & + SBCB + FCN          & 80.6 & +1.1 \\
                                & + SBCB + SegFix       & 81.0 & +1.5 \\
                                & + SBCB + FCN + SegFix & 81.2 & +1.7 \\
  \end{tabular}
  \label{table:ablation_segfix}
  \vspace{-1.2em}  
\end{table}

\subsection{Comparisons with SegFix}
\label{sec:ablation_segfix}

In Table \ref{table:ablation_segfix}, we compare our framework with SegFix \cite{yuan2020segfix}, a popular post-processing method.
We obtained the results for SegFix by using the open-source code, which refines the output prediction based on the offsets learned using HRNet2x.
Comparing the methods side-by-side, models trained with the SBCB framework, SegFix performs around $0.1\% \sim 0.4\%$ better than SBCB.
However, the SBCB combined with FCN (as mentioned in Section \ref{sec:ablation_auxiliary_signals}) results in competitive performance, significantly outperforming SegFix on two models.

Considering that SegFix is an independent post-processing model, our framework produces competitive results without any post-processing and additional parameters during inference.
Whereas, SegFix adds a post-processing module that requires separate training.
Also, motivated by the difficulty in prediction labels around the mask boundaries, SegFix is aimed to correct the predictions around the boundaries.
Therefore, the base model does not actively learn boundary-aware features.
On the other hand, our training framework conditions the backbone to be boundary-aware by solving SBD, as we see in Section \ref{sec:ablation_fscore}.
In other words, SegFix and our framework are complementary because boundary-aware predictions are easier for SegFix to correct.
This is evident by the major improvements of using SBCB along with SegFix, as shown in the table.

\begin{table}[ht!]
  \caption{Comparisons between DeepLabV3+ and GSCNN in the Cityscapes validation split. Note that the SBCB framework can be applied to train GSCNN.}
  \scriptsize
  \setlength\tabcolsep{4pt}
  \centering
  \begin{tabular}{c|c||cc}
    \multicolumn{2}{c||}{Model} & mIoU & $\Delta$ \\
  \hline\hline
    \multirow{3}{*}{DeepLabV3+} &                     & 79.5 & \\
                                & +SBCB (CASENet)     & 80.2 & +0.7 \\
                                & +SBCB (DDS)         & 80.6 & +1.1 \\
  \cdashline{2-4}\rule{0pt}{2.5ex} 
    \multirow{4}{*}{GSCNN}      &                  & 80.5 & +1.0 \\
                                & +Canny           & 80.6 & +1.1 \\
                                & SBD              & 80.0 & +0.5 \\
                                & +SBCB (CASENet)  & 80.9 & +1.4 \\
  \end{tabular}
  \label{table:ablation_gscnn}
  \vspace{-1.2em}  
\end{table}


\begin{table}[hb!]
  \caption{This table shows the configurations of the two common types of modifications on the ResNet backbone. Note that the outputs feature resolutions are in the order of Stem, Stages 1, Stage 2, Stage 3, and Stage 4.}
  \label{table:backbone_trick_config}
  \scriptsize
  \setlength\tabcolsep{4pt}
  \centering
  \begin{tabular}{c|cccc}
    Task & Stem Stride & Strides & Dilations & Resolutions \\
    \hline
    Original              & 2 & (1, 2, 2, 2) & (1, 1, 1, 1) & (1/2, 1/4, 1/8, 1/16, 1/32) \\
    Segmentation          & 2 & (1, 2, 1, 1) & (1, 1, 2, 4) & (1/2, 1/4, 1/8, 1/8, 1/8) \\
    Edge Det.             & 1 & (1, 2, 2, 1) & (2, 2, 2, 4) & (1, 1/2, 1/4, 1/8, 1/8) \\
  \end{tabular}
  \vspace{-1.2em}  
\end{table}

\begin{table*}[ht]
  \caption{Ablation studies of the ``Backbone Trick''. We modified the ResNet-101 backbone's stride and dilation at each stage to ensure the number of parameters is the same but generates larger feature maps. The paper \cite{xie2015hed} introduced this technique, which we prepend ``HED'' for backbone that uses this trick.}
  \begin{subtable}[t]{0.42\textwidth}
    \caption{Results on the Cityscapes validation split.}
    \scriptsize
    \setlength\tabcolsep{4pt}
    \centering
    \begin{tabular}{c||cc|cc}
    Head & mIoU & mF (ODS) & Param. & GFLOPs \\
    \hline\hline
    DeepLabV3+          & 79.8 & - & 60.2M   & 506  \\
    \hline
    CASENet             & \multirow{3}{*}{-} & 68.6 & 42.5M  & 417   \\
    DFF                 & & 70.0 & 42.8M  & 455    \\
    DDS                 & & 76.3 & 243.3M & 2661  \\
    \hline
    SBCB (CASENet)      & \textbf{81.0} & 75.1 & 60.2M  & 508 \\
    SBCB (DFF)          & 80.8 & 75.4 & 60.5M  & 545 \\
    SBCB (DDS)          & 80.8 & \textbf{76.5} & 261.0M & 2228 \\
    \end{tabular}
    \label{table:ablation_backbone_trick}
  \end{subtable}
  \vspace{\fill}
  \begin{subtable}[t]{0.28\textwidth}
    \flushright
    \caption{Results on the BDD100K validation split.}
    \scriptsize
    \setlength\tabcolsep{4pt}
    \centering
    \begin{tabular}{c||cc}
      Head & mIoU & mF (ODS) \\
    \hline\hline
      DeepLabV3+     & 59.8 & -  \\
    \hline
      CASENet        & \multirow{3}{*}{-} & 56.6 \\
      DFF            &                    & 58.1 \\
      DDS            &                    & 60.1 \\
    \hline
      SBCB (CASENet) & 62.4 & 59.3 \\
      SBCB (DFF)     & 62.0 & 58.9 \\
      SBCB (DDS)     & \textbf{63.5} & \textbf{60.5} \\
    \end{tabular}
    \label{table:ablation_hedresnet101_bdd100k}
  \end{subtable}
  \vspace{\fill}
  \begin{subtable}[t]{0.28\textwidth}
    \flushright
    \caption{Results on the Synthia dataset.}
    \scriptsize
    \setlength\tabcolsep{4pt}
    \centering
    \begin{tabular}{c||cc}
      Head & mIoU & mF (ODS) \\
    \hline\hline
      DeepLabV3+     & 77.0 & -  \\
    \hline
      CASENet        & \multirow{3}{*}{-} & 64.0 \\
      DFF            &                    & 65.6 \\
      DDS            &                    & 68.5 \\
    \hline
      SBCB (CASENet) & 78.0 & 67.5 \\
      SBCB (DFF)     & 77.8 & \textbf{68.9} \\
      SBCB (DDS)     & \textbf{78.6} & 68.4 \\
    \end{tabular}
    \label{table:ablation_hedresnet101_synthia}
  \end{subtable}
  \vspace{-1.2em}  
\end{table*}

\subsection{Comparisons with GSCNN}
\label{sec:ablation_gscnn}

GSCNN \cite{takikawa2019gscnn} is a popular semantic segmentation model with binary boundary detection multi-task architecture with a dedicated shape stream that branches out from the side layers similar to the SBD heads in the SBCB framework.
The key difference is that the features from the shape stream are explicitly merged into the semantic segmentation head.
GSCNN for ResNet-101 backbone is a customized DeepLabV3+ that uses an ASPP module.

It is difficult to compare apples to apples since loss functions, and we do not explicitly merge the features obtained in the SBD head to the segmentation head.
However, we will compare how well the SBCB framework can improve DeepLabV3+ against some of the configurations for GSCNN in Table \ref{table:ablation_gscnn}.
The baseline GSCNN is GSCNN without the image gradient (Canny Edge).
We also include the original configuration with Canny Edge denoted by ``+Canny''.
We also experimented with supervising the shape stream using the SBD task denoted by ``SBD'' and modified the shape stream by increasing the channels.
Finally, we used the SBCB framework on GSCNN denoted by ``+SBCB,'' which adds the SBD head on the backbone without any other modifications.

Compared with DeepLabV3+, GSCNN significantly improves by an additional $+1.0\%$.
Although lower than being supervised with binary boundaries, SBD supervision improves DeepLabV3+ by $+0.5\%$, proving that boundary signals can significantly improve semantic segmentation.
The SBCB framework significantly improves DeepLabV3+ by adding $0.7\%$ and $1.1\%$ with CASENet and DDS, respectively.
This also matches the improvements using the original GSCNN configuration.
Since the SBCB framework is flexible, it can be easily applied to GSCNN, giving an even higher improvement of $+1.4\%$.

\begin{table}[ht!]
  \caption{Comparison with SBD models on the Cityscapes validation split using the instance-sensitive ``thin'' evaluation protocol. $\dagger$: performance reported in \cite{liu2022dds}.}
  \scriptsize
  \setlength\tabcolsep{4pt}
  \centering
  \begin{tabular}{cc|c}
  Method & Backbone & mF (ODS) \\
  \hline\hline
  CASENet$\dagger$ & HED ResNet-101 & 68.1 \\
  SEAL$\dagger$    & HED ResNet-101 & 69.1 \\
  STEAL$\dagger$   & HED ResNet-101 & 69.7 \\
  DDS$\dagger$     & HED ResNet-101 & 73.8 \\
  \cdashline{1-3}\rule{0pt}{2.5ex}
  CSEL\cite{yu2021coupled} & HED ResNet-101 & 78.1 \\
  \hline
  DeepLabV3+ + SBCB (CASENet) & ResNet-101     & 77.8 \\
  DeepLabV3+ + SBCB (CASENet) & HED ResNet-101 & 78.4 \\
  DeepLabV3+ + SBCB (DDS)     & ResNet-101     & \textbf{78.8} \\
  DeepLabV3+ + SBCB (DDS)     & HED ResNet-101 & \textbf{78.8} \\
  \end{tabular}
  \label{table:ablation_sbd}
  \vspace{-1.2em}  
\end{table}

\begin{table}[ht!]
  \caption{Results to compare the boundary Fscore, evaluated on the Cityscapes validation split, between the baseline models and the use of SBCB training. The models are trained using the same hyperparameters where the backbones are set to the ResNet-101 backbone.}
  \scriptsize
  \setlength\tabcolsep{4pt}
  \centering
  \begin{tabular}{cc||cc|cc|cc|cc}
    Head & SBCB & 12px & $\Delta$ & 9px & $\Delta$ & 5px & $\Delta$ & 3px & $\Delta$ \\
  \hline\hline
    \multirow{2}{*}{PSPNet}     &            & 80.9 &      & 79.6 &      & 75.7 &      & 70.2 &      \\
                                & \checkmark & 83.3 & +2.4 & 82.1 & +2.5 & 78.5 & +2.8 & 73.3 & +3.1 \\
    \hline
    \multirow{2}{*}{DeepLabV3}  &            & 81.8 &      & 80.6 &      & 76.7 &      & 71.2 &      \\
                                & \checkmark & 83.4 & +1.6 & 82.2 & +1.6 & 78.7 & +2.0 & 73.4 & +2.2 \\
    \hline
    \multirow{2}{*}{DeepLabV3+} &            & 81.2 &      & 80.0 &      & 76.4 &      & 71.4 &      \\
                                & \checkmark & 83.0 & +1.8 & 81.8 & +1.8 & 78.5 & +2.1 & 73.7 & +2.3 \\
  \end{tabular}
  \label{table:ablation_fscore}
  \vspace{-1.2em}  
\end{table}

\subsection{Backbone Trick}
\label{sec:albation_backbone_trick}

In this section, we investigate the use of the ``backbone trick''.
In edge detection and SBD, we often use a modified backbone to increase the output resolutions of the stages without changing the number of parameters by modifying the strides and dilations for each stage.
The increase in resolution is necessary for edge detection as the edges are often small, and the feature maps need to be large enough to capture the edges.
Backbones such as ResNet were made for image classification and produced small feature maps unsuitable for edge detection.
It is also necessary not to change the number of parameters, as we want to use the pre-trained weights.
In semantic segmentation, we apply similar tricks to change the strides and dilations of the last two stages to retain the final feature resolution to $1/8$ of the input image size.
We show the common modifications for the ResNet backbone in Table \ref{table:backbone_trick_config}.

In Tables \ref{table:ablation_backbone_trick}, \ref{table:ablation_hedresnet101_bdd100k}, and \ref{table:ablation_hedresnet101_synthia}, we show results of using the HED version of ResNet-101 (HED ResNet-101) on Cityscapes, BDD100K and Synthia respectively.
Compared with the normal segmentation ResNet-101 in Table \ref{table:ablation_which_head}, the results are generally better for single-task as well as models trained with the SBCB framework.
Higher performance gains are seen in the Synthia dataset, where higher-resolution feature maps may benefit the detailed and precise ground truths.

Although the ``backbone trick'' is common for ResNet-101, it can be applied to other backbones, such as transformer backbones, as seen in Section \ref{sec:more_modern}.
Since the backbones are conditioned with SBD, the combination of SBD and the ``backbone trick'' can provide significant improvements without complex modeling.

\subsection{Does SBCB also improve SBD metrics?}
\label{sec:ablation_sbd}

Based on the previous ablations studies, it is clear that the SBCB framework improves the metrics for semantic segmentation.
We also evaluate the models trained using the SBCB framework on semantic boundary detection (SBD) performance as shown in Table \ref{table:ablation_sbd}.
We compare our DeepLabV3+ trained on the SBCB framework with state-of-the-art (SOTA) SBD models and CSEL, a SOTA joint semantic segmentation and semantic boundary detection model.
The table shows that those models trained on the SBCB framework can significantly outperform the SOTA single-task methods by $5\%$ to over $10\%$.
On joint modeling, our method can outperform CSEL without explicitly modeling in the semantic boundary detection head.
We aimed to condition the backbone for semantic segmentation, but the SBCB framework also improves the SBD performance due to being conditioned on semantic segmentation, which proves the effectiveness of the SBCB framework.

\subsection{Does SBCB improve segmentation around boundaries?}
\label{sec:ablation_fscore}

The SBCB framework improves segmentation quality around the mask boundaries.
In Table \ref{table:ablation_fscore}, we show boundary Fscores for baseline models and models trained on the SBCB framework.
The models trained using the SBCB framework constantly exhibit better boundary Fscores, especially when the trimap widths are smaller.
This means that conditioned backbones produce better segmentation quality around the mask's boundaries.

\begin{table*}[ht!]
  \caption{Effect of using SBCB for different CNN-based backbones.}
  \begin{subtable}[t]{0.48\textwidth}
    \caption{Results for the Cityscapes validation split.}
    \scriptsize
    \setlength\tabcolsep{4pt}
    \centering
    \begin{tabular}{cc|c||cc|cc}
      Head & Backbone & SBCB & mIoU & $\Delta$ & Fscore & $\Delta$ \\
    \hline\hline
      \multirow{2}{*}{DenseASPP}      & \multirow{2}{*}{ResNet-50}      &            & 77.5 &      & 69.0 & \\
                                      &                                 & \checkmark & 78.3 & +0.8 & 70.6 & +1.6 \\\hline
      \multirow{2}{*}{DenseASPP}      & \multirow{2}{*}{DenseNet-169}   &            & 76.6 &      & 69.0 & \\
                                      &                                 & \checkmark & 78.2 & +1.6 & 72.1 & +3.1 \\\hline
      \multirow{2}{*}{ASPP}           & \multirow{2}{*}{ResNeSt-101}    &            & 79.5 &      & 72.3 & \\
                                      &                                 & \checkmark & 80.3 & +0.8 & 75.2 & +2.9 \\\hline
      \multirow{2}{*}{OCR}            & \multirow{2}{*}{HR18}           &            & 78.9 &      & 71.9 & \\
                                      &                                 & \checkmark & 79.7 & +0.8 & 74.0 & +2.1 \\\hline
      \multirow{2}{*}{OCR}            & \multirow{2}{*}{HR48}           &            & 80.7 &      & 74.4 & \\
                                      &                                 & \checkmark & 82.0 & +1.3 & 77.7 & +3.7 \\\hline
      \multirow{2}{*}{ASPP}           & \multirow{2}{*}{MobileNetV2}    &            & 73.9 &      & 66.2 & \\
                                      &                                 & \checkmark & 74.4 & +0.5 & 68.3 & +2.1 \\\hline
      \multirow{2}{*}{LRASPP}         & \multirow{2}{*}{MobileNetV3}    &            & 64.5 &      & 58.0 & \\
                                      &                                 & \checkmark & 67.5 & +3.0 & 62.1 & +4.1 \\
    \end{tabular}
    \label{table:evaluation_backbones_cityscapes}
  \end{subtable}
  \vspace{\fill}
  \begin{subtable}[t]{0.48\textwidth}
    \flushright
    \caption{Results for the Synthia Datast.}
    \scriptsize
    \setlength\tabcolsep{4pt}
    \centering
    \begin{tabular}{cc|c||cc}
      Head & Backbone & SBCB & mIoU & $\Delta$ \\
    \hline\hline
      \multirow{2}{*}{DenseASPP}      & \multirow{2}{*}{ResNet-50}      &            & 69.6 &      \\
                                      &                                 & \checkmark & 70.5 & +0.9 \\\hline
      \multirow{2}{*}{DenseASPP}      & \multirow{2}{*}{DenseNet-169}   &            & 71.3 &      \\
                                      &                                 & \checkmark & 72.0 & +0.7 \\\hline
      \multirow{2}{*}{ASPP}           & \multirow{2}{*}{ResNeSt-101}    &            & 72.3 &      \\
                                      &                                 & \checkmark & 73.8 & +1.5 \\\hline
      \multirow{2}{*}{OCR}            & \multirow{2}{*}{HR18}           &            & 70.1 &      \\
                                      &                                 & \checkmark & 70.9 & +0.8 \\\hline
      \multirow{2}{*}{OCR}            & \multirow{2}{*}{HR48}           &            & 74.3 &      \\
                                      &                                 & \checkmark & 76.0 & +1.7 \\\hline
      \multirow{2}{*}{ASPP}           & \multirow{2}{*}{MobileNetV2}    &            & 65.3 &      \\
                                      &                                 & \checkmark & 67.0 & +1.7 \\\hline
      \multirow{2}{*}{LRASPP}         & \multirow{2}{*}{MobileNetV3}    &            & 60.8 &      \\
                                      &                                 & \checkmark & 64.8 & +4.0 \\
    \end{tabular}
    \label{table:evaluation_backbones_synthia}
  \end{subtable}
\end{table*}

\begin{table*}[ht!]
  \caption{Effect of using SBCB on different segmentation heads. Note that the backbones for all models are set to ResNet-101.}
  \begin{subtable}[t]{0.48\textwidth}
    \caption{Results on the Cityscapes validation split.}
    \scriptsize
    \setlength\tabcolsep{4pt}
    \centering
    \begin{tabular}{c|c||cc|cc}
      Head & SBCB & mIoU & $\Delta$ & Fscore & $\Delta$ \\
    \hline\hline
      \multirow{2}{*}{FCN}            &            & 74.6 &      & 69.3 & \\
                                      & \checkmark & 76.3 & +1.7 & 71.6 & +2.3\\\hline
      \multirow{2}{*}{PSPNet}         &            & 77.6 &      & 70.2 & \\
                                      & \checkmark & 78.7 & +1.1 & 73.2 & +3.0 \\\hline
      \multirow{2}{*}{ANN}            &            & 77.4 &      & 70.1 & \\
                                      & \checkmark & 79.0 & +1.6 & 72.8 & +2.7 \\\hline
      \multirow{2}{*}{GCNet}          &            & 77.8 &      & 70.2 & \\
                                      & \checkmark & 78.9 & +1.1 & 73.0 & +2.8 \\\hline
      \multirow{2}{*}{ASPP}           &            & 79.2 &      & 71.2 & \\
                                      & \checkmark & 79.9 & +0.7 & 73.4 & +2.2 \\\hline
      \multirow{2}{*}{DNLNet}         &            & 78.7 &      & 71.2 & \\
                                      & \checkmark & 79.7 & +1.0 & 73.6 & +2.4 \\\hline
      \multirow{2}{*}{CCNet}          &            & 79.2 &      & 71.9 & \\
                                      & \checkmark & 80.1 & +0.9 & 73.9 & +2.0 \\\hline
      \multirow{2}{*}{UPerNet}        &            & 78.1 &      & 71.9 & \\
                                      & \checkmark & 78.9 & +0.8 & 73.9 & +2.0 \\\hline
      \multirow{2}{*}{OCR}            &            & 78.2 &      & 70.6 & \\
                                      & \checkmark & 80.2 & +2.0 & 74.4 & +3.8 \\
    \end{tabular}
    \label{table:evaluation_heads_cityscapes}
  \end{subtable}
  \vspace{\fill}
  \begin{subtable}[t]{0.48\textwidth}
    \flushright
    \caption{Results on the Synthia Dataset.}
    \scriptsize
    \setlength\tabcolsep{4pt}
    \centering
    \begin{tabular}{c|c||cc}
      Head & SBCB & mIoU & $\Delta$ \\
    \hline\hline
      \multirow{2}{*}{FCN}            &            & 70.0 &      \\
                                      & \checkmark & 70.9 & +0.9 \\\hline
      \multirow{2}{*}{PSPNet}         &            & 70.5 &      \\
                                      & \checkmark & 71.7 & +1.2 \\\hline
      \multirow{2}{*}{ANN}            &            & 70.4 &      \\
                                      & \checkmark & 71.8 & +1.4 \\\hline
      \multirow{2}{*}{GCNet}          &            & 70.8 &      \\
                                      & \checkmark & 71.4 & +0.6 \\\hline
      \multirow{2}{*}{ASPP}           &            & 70.9 &      \\
                                      & \checkmark & 71.9 & +1.0 \\\hline
      \multirow{2}{*}{DNLNet}         &            & 70.5 &      \\
                                      & \checkmark & 71.9 & +1.4 \\\hline
      \multirow{2}{*}{CCNet}          &            & 70.8 &      \\
                                      & \checkmark & 71.3 & +0.5 \\\hline
      \multirow{2}{*}{UPerNet}        &            & 72.4 &      \\
                                      & \checkmark & 73.1 & +0.7 \\\hline
      \multirow{2}{*}{OCR}            &            & 69.7 &      \\
                                      & \checkmark & 72.4 & +2.7 \\
    \end{tabular}
    \label{table:evaluation_heads_synthia}
  \end{subtable}
  \vspace{-1.2em}  
\end{table*}

\section{Applications of SBCB}
\label{sec:applications_sbcb}

In this section, we focus on the applications of the SBCB framework.
In Sections \ref{sec:applications_backbones} and \ref{sec:applications_heads}, we show the effectiveness of applying SBCB training on a broad range of backbones and popular segmentation heads.
In Section \ref{sec:applications_cityscapes}, we benchmark our method of applying SBCB for DeepLabV3+ on the Cityscapes dataset and compare our results with the state-of-the-art (SOTA) methods.

\subsection{Different Backbones}
\label{sec:applications_backbones}

In Tables \ref{table:evaluation_backbones_cityscapes} and \ref{table:evaluation_backbones_synthia}, we show the improvements when models are trained with the SBCB framework on several backbones.
We use two datasets with varying degrees of annotation qualities to show the robustness and consistency of the SBCB framework.
The two tables show that the SBCB framework consistently and significantly improves the IoU even when the backbones differ.
Note that the backbones evaluated here are mature ConvNet architectures, but we will explore the effects of SBCB on customized backbones and modern methods like the ConvNeXt and SegFormer in Sections \ref{sec:applications}.
We provide qualitative results of the SBCB framework on the Cityscapes dataset in Section \ref{sec:appendix_qualitative}.

\subsection{Different Heads}
\label{sec:applications_heads}

In Tables \ref{table:evaluation_heads_cityscapes} and \ref{table:evaluation_heads_synthia}, we show the performances of models trained using the SBCB framework for different heads.
Note that the backbone for the models is set to ResNet-101.
The tables show that the SBCB framework consistently improves the IoU and boundary Fscore for various segmentation heads.
We provide qualitative results of the SBCB framework on the Cityscapes dataset in Section \ref{sec:appendix_qualitative}.

\begin{table}[ht]
  \caption{Comparison of our method and state-of-the-art methods on the Cityscapes validation split. The methods are only trained with fine-annotation data and without additional coarse training data and Mapillary Vistas pre-training.}
  \scriptsize
  \setlength\tabcolsep{2.0pt}
  \centering
  \begin{tabular}{cc|c}
  Method & Backbone & mIoU \\
  \hline\hline
  PSPNet \cite{zhao2017pspnet}             & ResNet-101  & 78.8 \\
  DeepLabV3+ \cite{chen2018deeplabv3plus}  & ResNet-101  & 78.8 \\
  CCNet \cite{huang2019ccnet}              & ResNet-101  & 80.5 \\
  DANet \cite{fu2019danet}                 & ResNet-101  & 81.5 \\
  \hline
  GSCNN \cite{takikawa2019gscnn}           & ResNet-38   & 80.8 \\ 
  RPCNet \cite{zhen2020rpcnet}             & ResNet-101      & 82.1 \\
  CSEL \cite{yu2021coupled}                & HED ResNet-101  & 83.7 \\
  \hline
  DeepLabV3+ SBCB & ResNet-101     & 82.2 \\
  DeepLabV3+ SBCB & HED ResNet-101 & 82.6 \\
  \end{tabular}
  \label{table:cityscapes_val_split}
  \vspace{-1.2em}  
\end{table}

\begin{table}[ht]
  \caption{Comparison of our method and state-of-the-art methods on the Cityscapes test split. The methods are only trained with fine-annotation data and without additional coarse training data and Mapillary Vistas pre-training.}
  \scriptsize
  \setlength\tabcolsep{2.0pt}
  \centering
  \begin{tabular}{cc|c}
  Method & Backbone & mIoU \\
  \hline\hline
  PSPNet \cite{zhao2017pspnet}             & ResNet-101  & 78.4 \\
  PSANet \cite{zhao2018psanet}             & ResNet-101  & 80.1 \\
  SeENet \cite{pang2019seenet}             & ResNet-101  & 81.2 \\
  ANNNet \cite{zhu2019ann}                 & ResNet-101  & 81.3 \\
  CCNet \cite{huang2019ccnet}              & ResNet-101  & 81.4 \\
  DANet \cite{fu2019danet}                 & ResNet-101  & 81.5 \\
  \hline
  RPCNet \cite{zhen2020rpcnet}             & ResNet-101  & 81.8 \\
  CSEL \cite{yu2021coupled}                & HED ResNet-101  & 82.1 \\
  \hline
  DeepLabV3+ SBCB & ResNet-101 & 81.4 \\
  DeepLabV3+ SBCB & HED ResNet-101 & 81.0 \\
  \end{tabular}
  \label{table:cityscapes_test_split}
  \vspace{-1.2em}  
\end{table}

\subsection{Cityscapes Benchmarks}
\label{sec:applications_cityscapes}

\noindent\textbf{Cityscapes Validation Split.}
In Table \ref{table:cityscapes_val_split}, we show the performance of DeepLabV3+ trained with the SBCB framework and compare it to other SOTA models on the Cityscapes validation split.
The top group is SOTA single task methods, the middle group is SOTA joint task methods, and the last group is our baseline models with different backbones.
We can see that even without the recent strong heads, the SBCB framework enables these methods to outperform single-task methods and perform competitively with joint-task models.
Our method outperforms two popular multi-task methods, GSCNN and RPCNet, while using off-the-shelf segmentation head and backbone.

\noindent\textbf{Cityscapes Benchmark.}
In Table \ref{table:cityscapes_test_split}, we show the performance of DeepLabV3+ trained with the SBCB framework and compare it to other SOTA models on the Cityscapes Benchmark.
Although we could not achieve better results than SOTA multi-task methods, DeepLabV3+ trained with SBDB matched some SOTA methods and proved competitive.

\section{More Applications}
\label{sec:applications}

In this section, we present more applications for the SBCB framework.
In Section \ref{sec:more_ade20k}, we experiment on the challenging ADE20k dataset.
In Sections \ref{sec:more_bisenet} and \ref{sec:more_stdc}, we apply SBCB training on recent lightweight segmentation architectures and show the flexibility and effectiveness of the SBCB framework.
In Section \ref{sec:more_modern}, we applied SBCB training to ConvNeXt and Segformer.
Finally, in Section \ref{sec:more_explit_fusion}, we introduce methods of explicitly fusing the two heads and compare the methods to the proposed SBCB framework.

\begin{table}[ht]
  \caption{Results using ResNet backbones on the ADE20k validation split.}
  \scriptsize
  \setlength\tabcolsep{4pt}
  \centering
  \begin{tabular}{cccc||cc}
    Head & Backbone & Batch & SBCB  & mIoU & $\Delta$ \\
  \hline\hline
    \multirow{4}{*}{PSPNet} & 50  & 8 &            & 39.9 &  \\
                            & 50  & 8 & \checkmark & 40.6 & +0.7 \\
  \cdashline{2-6}\rule{0pt}{2.5ex} 
                            & 101 & 4 &            & 38.2 &  \\
                            & 101 & 4 & \checkmark & 38.7 & +0.5 \\
  \hline
    \multirow{4}{*}{DeepLabV3+} & 50  & 8 &            & 41.5 &  \\
                                & 50  & 8 & \checkmark & 42.0 & +0.5 \\
  \cdashline{2-6}\rule{0pt}{2.5ex} 
                                & 101 & 4 &            & 37.7 &  \\
                                & 101 & 4 & \checkmark & 38.2 & +0.5 \\
  \end{tabular}
  \label{table:applications_ade20k}
  \vspace{-1.2em}  
\end{table}

\subsection{Experiments on ADE20k}
\label{sec:more_ade20k}

We perform additional experiments on ADE20k which is another challenging dataset known for having 150 different classes \cite{Zhou2017ADE20k}.
We train DeepLabV3+ with ResNet-50 and ResNet-101 as the backbone and compared the results against ones trained using the SBCB framework.
The results show that the SBCB framework improves the base models by around $0.5\%$ which is shown in Table \ref{table:applications_ade20k}.

\begin{table}[ht]
  \caption{Results for BiSeNet and STDC on Cityscapes validation split.}
  \scriptsize
  \setlength\tabcolsep{4pt}
  \centering
  \begin{tabular}{cc||cc|cc}
    Model & SBCB & mIoU & $\Delta$ & Fscore & $\Delta$ \\
  \hline\hline
    \multirow{2}{*}{BiSeNetV1 R50} &            & 74.3 &      & 66.0 &  \\
                                   & \checkmark & 75.4 & +1.1 & 69.9 & +3.9 \\\hline
    \multirow{2}{*}{BiSeNetV2}     &            & 70.7 &      & 63.8 &  \\
                                   & \checkmark & 71.6 & +0.9 & 66.2 & +2.4 \\\hline
    STDC V1 FCN (+Detail Head)     &            & 73.7 &      & 66.5 &  \\
    STDC V1 FCN                    & \checkmark & 75.4 & +1.7 & 67.9 & +1.4 \\
  \end{tabular}
  \label{table:more_mobile}
  \vspace{-1.2em}  
\end{table}

\subsection{BiSeNet}
\label{sec:more_bisenet}

We applied the SBCB framework to Bilateral Segmentation Network (BiSeNet) V1 and V2, which are models specialized for real-time semantic segmentation \cite{Yu2018BiSeNet,Yu2020BiSeNetV2}.
In both versions, the backbone is split into two paths.
The Detail Path (or Spatial Path) is a shallow ConvNet composed of a few stages that retain large feature resolutions.
For BiSeNetV1, the number of stages is set to four, while it is set to three in BiSeNetV2.
On the other hand, the Semantic Path (or Context Path) is a deeper ConvNet designed to capture high-level semantics.
While in BiSeNetV1, the Semantic Path uses off-the-shelf architectures such as ResNet-50, BiSeNetV2 uses a customized six-stage ConvNet where the features from the middle stages are supervised using FCN auxiliary heads.

We applied the SBCB framework by choosing the stages (sides) of the backbone to be supervised by the SBD head.
We take three stages from the Detail Path for the Binary Sides for the SBD head and use the last stage of the Semantic Path for the Semantic Side.
Note that we do not modify the original model in any way; we only add the SBD head by taking the mid features of the backbones.
See Appendix \ref{sec:appendix_bisenet} for details.

The SBCB framework's results on BiSeNet (V1 and V2) are shown in Table \ref{table:more_mobile}.
As expected, using the SBCB framework improves the models in both IoU and boundary Fscore.
This proves the SBCB framework can apply to non-common architectures and expect performance gains.

\subsection{STDC}
\label{sec:more_stdc}

Like BiSeNet, the STDC network is efficient for real-time semantic segmentation \cite{Fan2021STDC}.
However, the STDC network is a single branch network that replaces the Detail Path with the Detail Head that uses the features from the third stage to perform ``detail guidance'' only during the training phase.
The Detail Head is supervised with ``Detail GT,'' which is generated using a multi-scale Laplacian Convolution kernel in an on-the-fly manner similar to our method.
The detail GT contains spatial details like boundaries and corners.

In this section, we replace the Detail Head with the SBD head and train using the SBCB framework.
We take the first four stages of the backbone for the Binary Sides and use the output of the FFM as the Semantic Side for the SBD head (see Appendix \ref{sec:appendix_stdc}).
The results are shown in Table \ref{table:more_mobile}, where we compare the original STDC with STDC that replaced the Detail Head with our SBD head.
We can see significant improvements in using SBD as the auxiliary task with substantial improvements in the IoU.
The Detail Head aimed at improving the segmentation quality around the boundaries, but our framework shows higher improvements in the boundary Fscore.

\begin{table}[ht]
  \caption{Results of the SBCB framework on modern backbones/architectures on the Cityscapes validation split.}
  \scriptsize
  \setlength\tabcolsep{4pt}
  \centering
  \begin{tabular}{cc|c||cc|cc}
    Head & Backbone & SBCB & mIoU & $\Delta$ & Fscore & $\Delta$ \\
  \hline\hline
    \multirow{3}{*}{UPerNet}        & \multirow{2}{*}{ConvNeXt-base}  &            & 81.8 &      & 74.4 & \\
                                    &                                 & \checkmark & 82.0 & +0.2 & 75.5 & +1.1 \\
                                    & Mod ConvNeXt-base               & \checkmark & 82.2 & +0.4 & 76.5 & +2.1\\\hline 
    \multirow{3}{*}{SegFormer}      & \multirow{2}{*}{MiT-b0}         &            & 75.5 &      & 66.9 & \\
                                    &                                 & \checkmark & 76.5 & +1.0 & 68.1 & +1.2 \\
                                    & Mod MIT-b0                      & \checkmark & 76.8 & +1.3 & 69.7 & +2.8 \\\hline
    \multirow{3}{*}{SegFormer}      & \multirow{2}{*}{MiT-b2}         &            & 80.9 &      & 73.2 & \\
                                    &                                 & \checkmark & 81.1 & +0.2 & 74.7 & +1.5 \\
                                    & Mod MIT-b2                      & \checkmark & 81.6 & +0.7 & 76.0 & +2.8 \\\hline
    \multirow{2}{*}{SegFormer}      & \multirow{2}{*}{MiT-b4}         &            & 81.6 &      & 75.5 & \\
                                    &                                 & \checkmark & 82.2 & +0.6 & 76.7 & +1.2 \\
  \end{tabular}
  \label{table:more_modern}
  \vspace{-1.2em}  
\end{table}

\subsection{ConvNeXt and SegFormer}
\label{sec:more_modern}

In this section, we applied the SBCB framework and the ``Backbone Trick'' to two modern architectures.
ConvNeXt is a backbone composed of pure ConvNet components with design elements borrowed from vision Transformers (ViT) \cite{dosovitskiy2021vit,Liu2022ConvNeXt}.
On the other hand, SegFormer is a full-blown segmentation architecture composed of a ViT-inspired backbone called the Mix Transformer (MiT), with a lightweight All-MLP segmentation head \cite{xie2021segformer}
Both architectures exhibit hierarchical feature extraction, which is compatible with the SBCB framework.
The results of applying the SBCB framework are shown in Table \ref{table:more_modern}.
We also compare the effects of adding the ``Backbone Trick'' denoted by ``Mod'' in the backbones.
From the table, we can see that the SBCB framework can still be applied to improve these modern architectures and provide consistent performance gains in both IoU and boundary Fscore.

\begin{table}[ht]
  \caption{Results using explicit feature fusion at the heads on the Cityscapes validation split.}
  \scriptsize
  \setlength\tabcolsep{4pt}
  \centering
  \begin{tabular}{c|c||cc}
    \multicolumn{2}{c||}{Model} & mIoU & $\Delta$ \\
  \hline\hline
  \multirow{3}{*}{PSPNet}
                                &            & 77.6 & \\
                                & +SBCB      & 78.7 & +1.1 \\
  \cdashline{2-4}\rule{0pt}{2.5ex} 
                                & Channel-Merge & \textbf{79.1} & +1.5 \\
  \hline
    \multirow{4}{*}{DeepLabV3+} &                     & 79.5 & \\
                                & +SBCB               & 80.2 & +0.7 \\
  \cdashline{2-4}\rule{0pt}{2.5ex} 
                                & Two-Stream Merge   & \textbf{80.5} & +1.0 \\
                                & Channel-Merge      & \textbf{80.5} & +1.0 \\
  \end{tabular}
  \label{table:more_fusion}
  \vspace{-1.2em}  
\end{table}

\subsection{Explicit Feature Fusion}
\label{sec:more_explit_fusion}

We provide two feature fusion techniques to utilize the features learned in the SBD head that can further be applied to improve segmentation.
The first technique uses simple channel concatenation with few convolutional layers to motivate feature fusion, called the Channel-Merge method.
Another technique is a naive merge used in GSCNN, where the features learned in the SBD head are also used in the ASPP head for DeepLabV3+, similar to GSCNN.
We call the latter method the Two-Stream Merge method.
The two fusion architectures are explained in more detail in Appendix \ref{sec:appendix_explicit_fusion}.

Table \ref{table:more_fusion} shows the results of two baseline architectures with the SBCB framework and feature fusion methods applied.
We can see that the feature fusion methods can further improve the segmentation performance.
It also comes with the downside of making the segmentation head dependent on the SBD head, which increases computational costs.
We believe that the SBCB framework helps boost existing segmentation models, and the SBD heads could further inspire exciting architectures for joint architectures like Channel-Merge and Two-Stream Merge.

\section{Conclusion}
\label{sec:conclusion}

We have proposed the SBCB framework, a simple yet effective training framework that boosts segmentation performance.
In the framework, a semantic boundary detection (SBD) head is applied to the hierarchical features of the backbone which is supervised by semantic boundaries.
We have explored different SBD heads for the SBCB framework and showed that the CASENet architecture significantly improves segmentation quality without adding many parameters during training.
Our experiments show that the SBCB framework improves segmentation quality on many popular backbones and segmentation heads.
It also improves the segmentation quality around the boundaries which was evaluated on boundary F-score.
We also have experimented with other customized backbones and recent transformer architectures to show that the SBCB framework is versatile.
Not only is the SBCB framework effective, but we have also provided modifications and methods of explicit feature fusion to promote the broader use of semantic boundaries for semantic segmentation.


\clearpage
\begin{appendices}

\section{On-the-fly Boundary Generation}
\label{sec:appendix_generation}

In this section, we will explain the on-the-fly (OTF) semantic boundary generation algorithm in detail.
For a single label $l$, we apply a signed distance function (SDF) on the inner and outer masks, where the inner mask represents the pixels that are $l$ and the outer mask represents pixels that are not $l$.
We can then take the sum of the inner and outer masks and use the pixels under the radius as the boundary pixels.
When instance segmentation maps are available, we generate per-instance distance maps, which we threshold using the same radius.
We sum all the boundaries for category $l$ (with instance boundaries) and binaries the resulting boundaries.
We repeat this step for every label until we have $L$ labels.
We concatenate the $L$ boundaries to form a $L \times H \times W$ semantic boundary tensor.

\begin{figure}[ht!]
  \centering
  \includegraphics[width=0.5\linewidth]{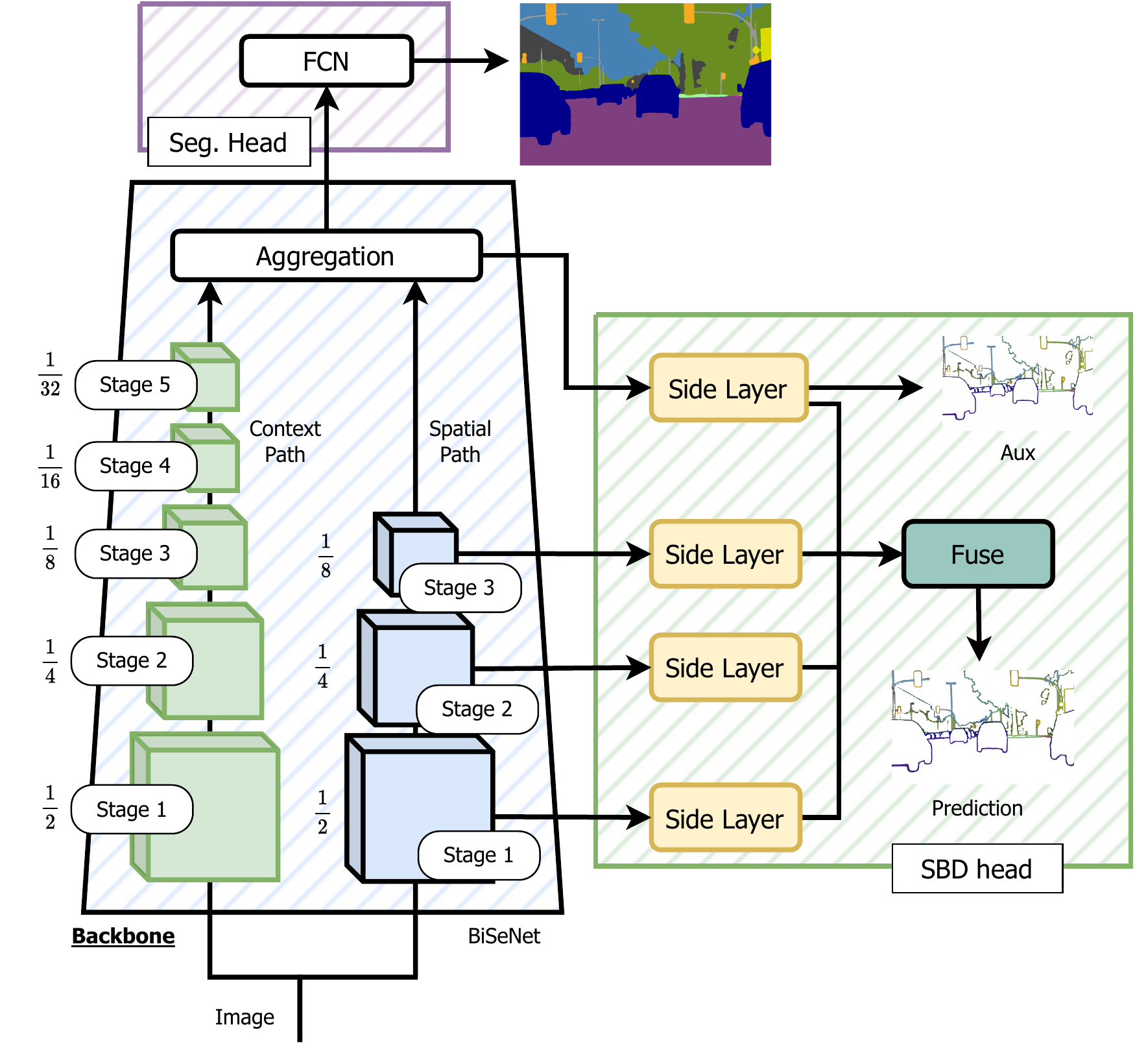}
  \caption{Architecture diagram for applying the SBCB framework to BiSeNet.}
  \label{fig:bisenet_sbcb_diagram}
  \vspace{-1.5em}  
\end{figure}

\begin{figure}[ht!]
  \centering
  \includegraphics[width=0.5\linewidth]{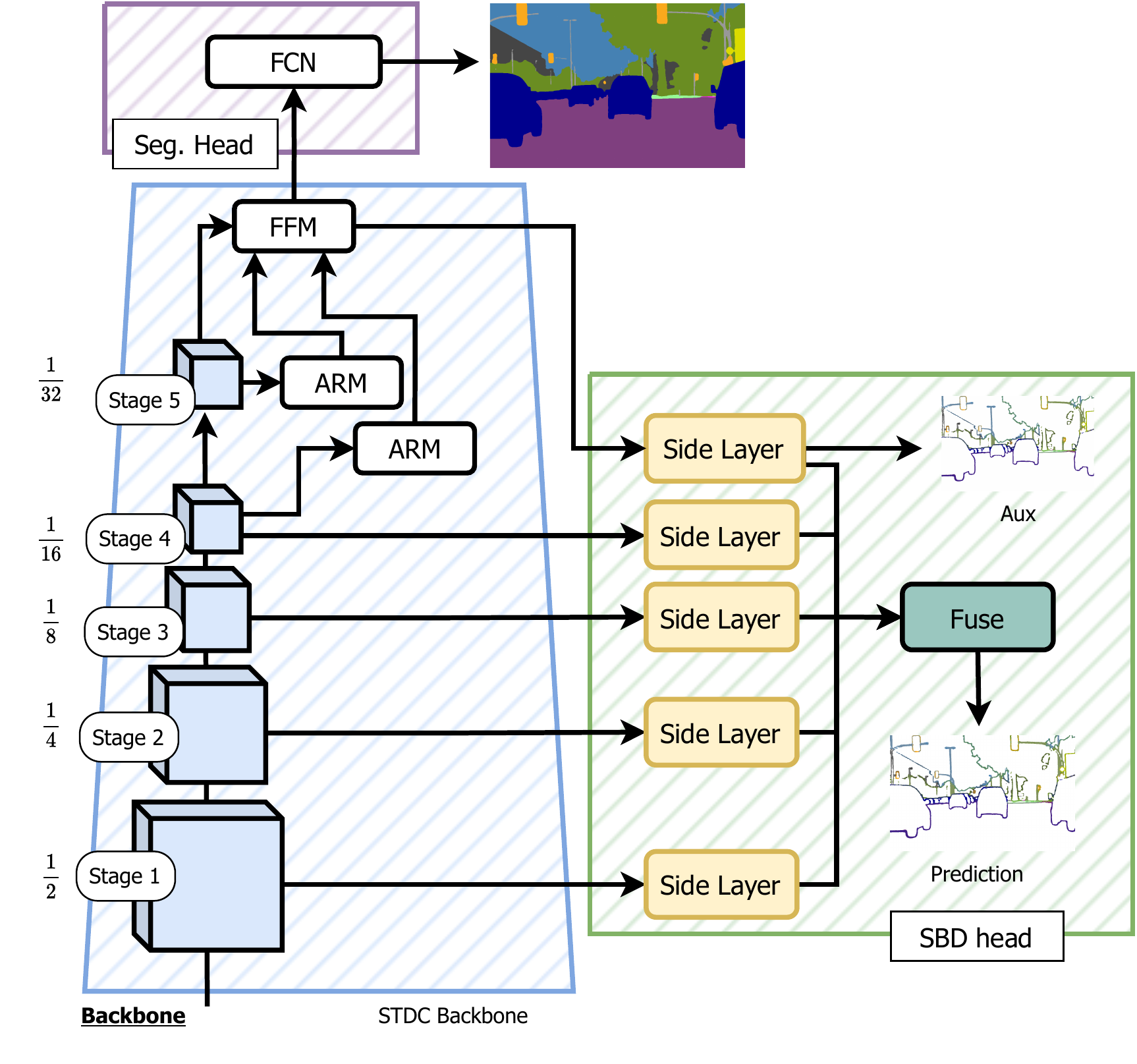}
  \caption{Architecture diagram for applying the SBCB framework to STDC backbone.}
  \label{fig:stdc_sbcb_diagram}
  \vspace{-1.5em}  
\end{figure}

\begin{figure}[ht!]
  \centering
  \includegraphics[width=0.5\linewidth]{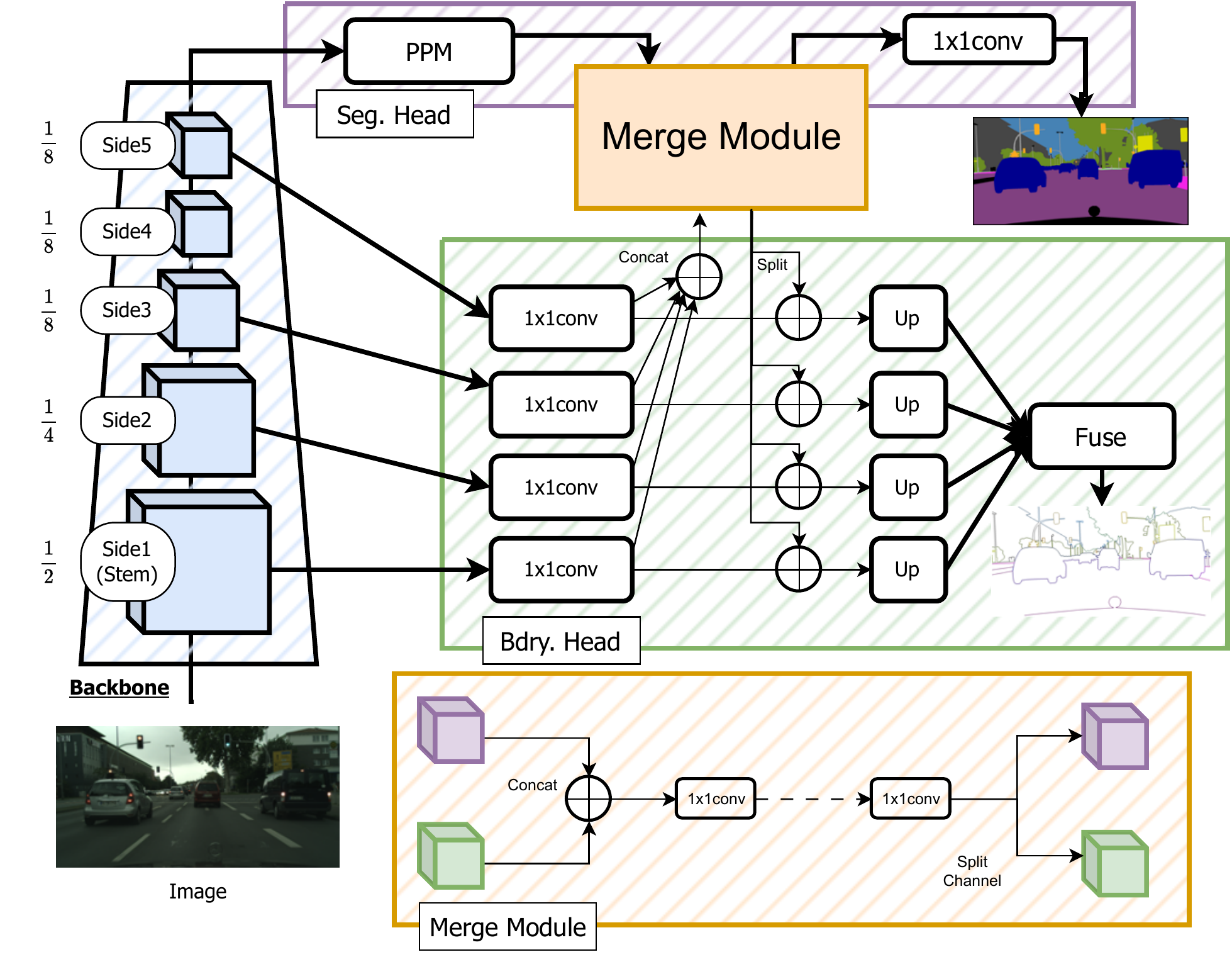}
  \caption{Diagram of applying Channel-Merge module for explicit feature fusion based on the SBCB framework.}
  \label{fig:ppm_channel_merge_sbcb_diagram}
  \vspace{-1.5em}  
\end{figure}

\begin{figure}[ht!]
  \centering
  \includegraphics[width=0.5\linewidth]{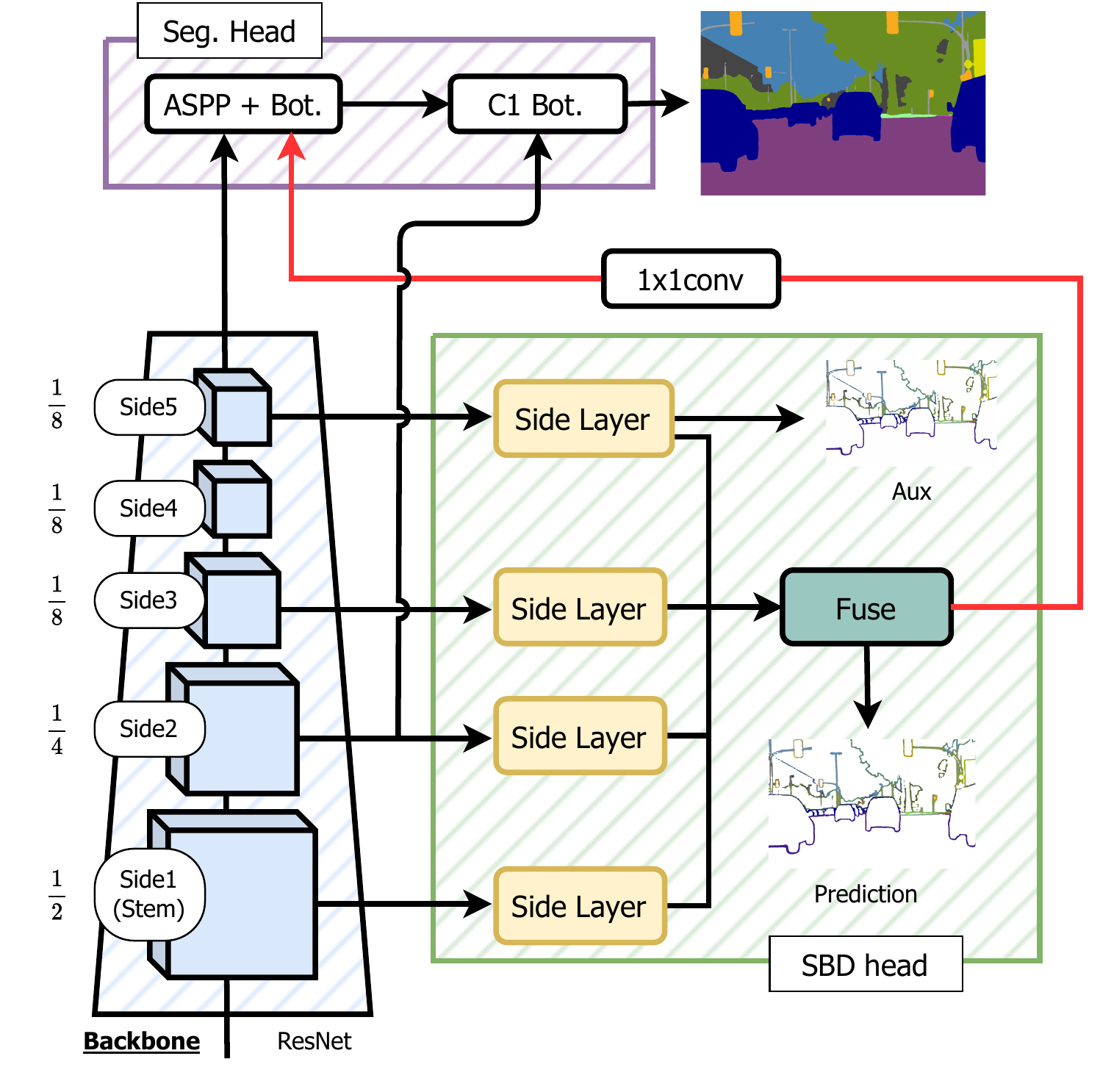}
  \caption{Diagram of applying two-stream approach for explicit feature fusion in the SBCB framework. The architecture is modeled after the GSCNN architecture.}
  \label{fig:twostream_sbcb_diagram}
  \vspace{-1.5em}  
\end{figure}

\section{Qualitative Visualizations}
\label{sec:appendix_qualitative}

\begin{figure*}[!p]
  \centering
  \includegraphics[width=\textwidth]{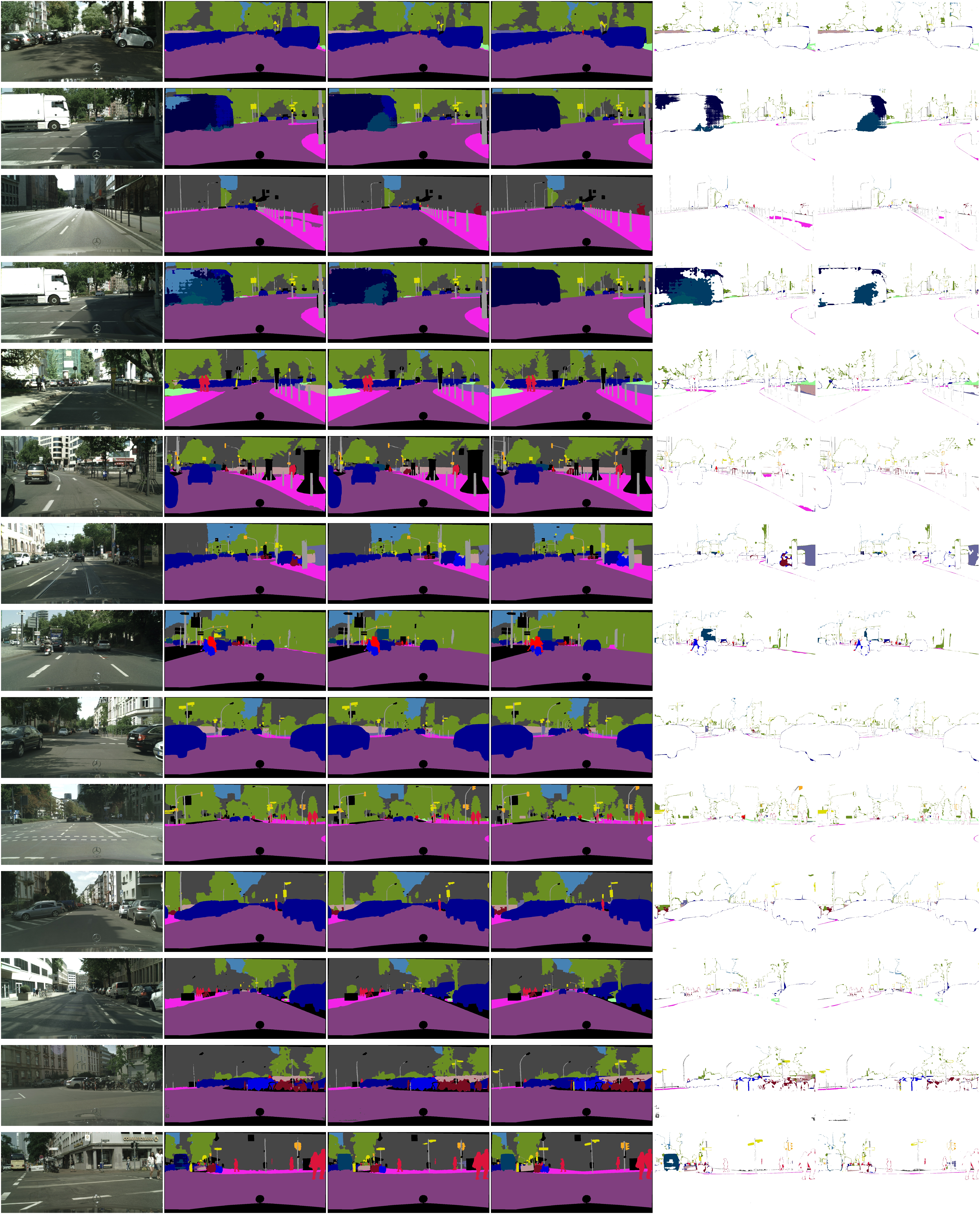}
  \caption{Visualization of the segmentation masks and segmentation errors for the models in Table \ref{table:evaluation_backbones_cityscapes}. Each column, from the left, represents the input image, prediction without SBCB, prediction with SBCB, ground truth, segmentation errors without SBCB, and segmentation error with SBCB. We visualize two samples (two rows) per backbone. From the top row, the backbones are ResNet-50, DenseNet-169, ResNeSt-101, HR18, HR48, MobileNetV2, and MobileNetV3. Best seen in color and zoomed in.}
  \label{fig:diff_back_qualitative}
\end{figure*}

\begin{figure*}[!p]
  \centering
  \includegraphics[width=\textwidth]{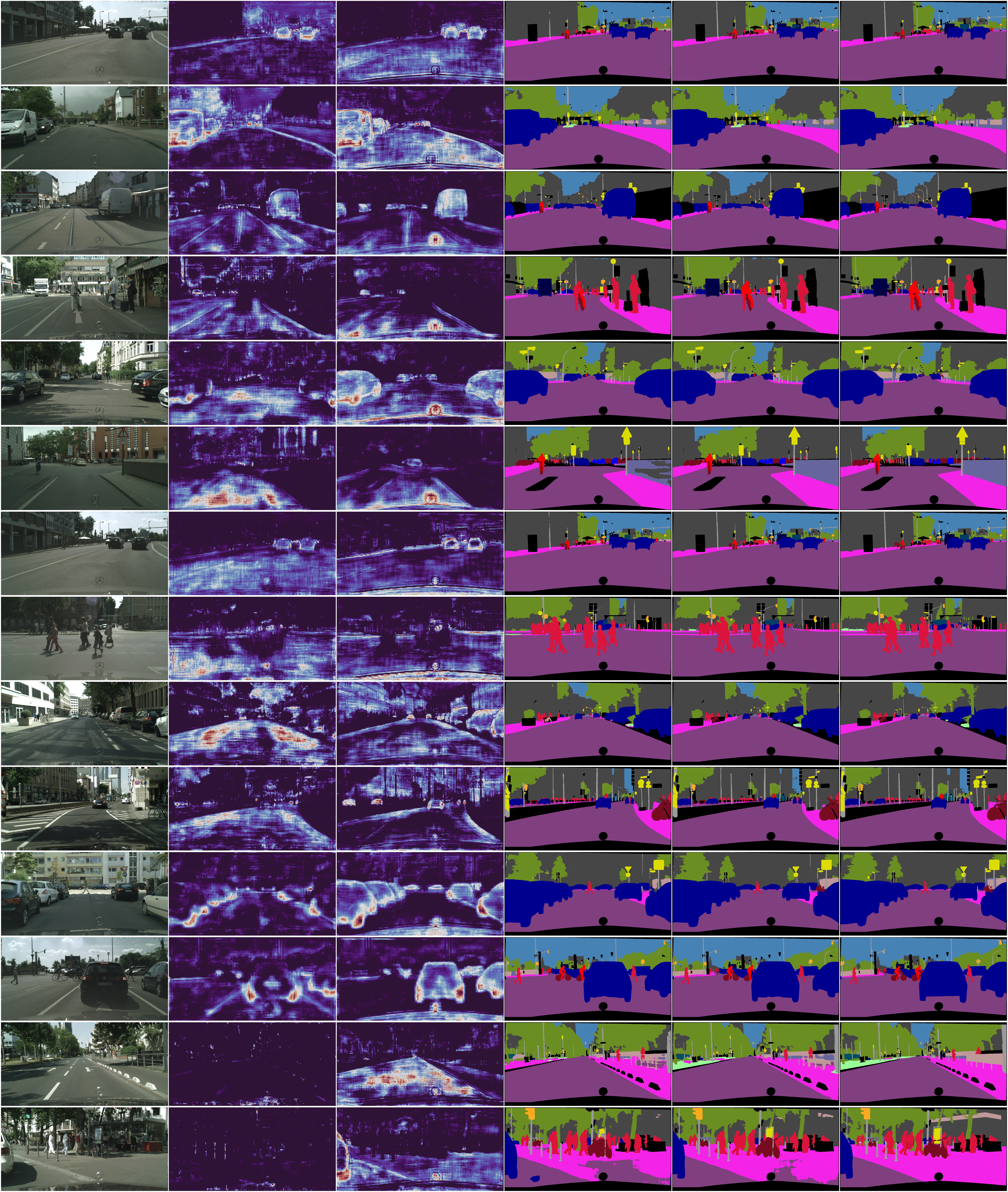}
  \caption{Visualization of the segmentation masks and segmentation errors for the models in Table \ref{table:evaluation_heads_cityscapes}. Each column, from the left, represents the input image, backbone feature without SBCB, backbone feature with SBCB, prediction without SBCB, prediction with SBCB, and ground truth. We visualize two samples (two rows) per segmentation head. From the top row, the segmentation heads are FCN, ANN, GCNet, DNLNet, CCNet, UperNet, and OCR. Best seen in color and zoomed in.}
  \label{fig:diff_head_qualitative}
\end{figure*}

We show qualitative visualizations for the results in Tables \ref{table:evaluation_backbones_cityscapes} and \ref{table:evaluation_heads_cityscapes} in Figures \ref{fig:diff_back_qualitative} and \ref{fig:diff_head_qualitative} respectively.

\section{BiSeNet + SBCB}
\label{sec:appendix_bisenet}

In Figure \ref{fig:bisenet_sbcb_diagram}, we show a detailed architecture diagram showing which features of the BiSeNet backbone are used in the SBD head.
In both BiSeNet V1 and V2, the architecture is composed of a Context Path and a Spatial Path.
We use the three stages of the spatial path for the earlier Side Layers of the SBD head.
We used the last feature of the Aggregation Layer for the last Side Layer.

\section{STDC + SBCB}
\label{sec:appendix_stdc}

In Figure \ref{fig:stdc_sbcb_diagram}, we show a detailed architecture diagram showing how we applied the SBCB framework to the STDC architecture.
The architecture is more reminiscent of a ResNet-like hierarchical backbone, but the original STDC applies a Detail Head, which uses the features of the third stage.
Instead, we remove the Detail Head and instead add an SBD head by using the first four stages for the binary side layer and the final output of the FFM as the input to the semantic side layer.

\section{Explicit Feature Fusion Architectures}
\label{sec:appendix_explicit_fusion}

In Figure \ref{fig:ppm_channel_merge_sbcb_diagram}, we show the proposed explicit feature fusion architecture built on top of the SBCB framework called the Channel-Merge module.
The diagram shows a backbone with hierarchical features and a PPM head used in the PSPNet.
The Channel-Merge module uses the features before upsampling in the Side Layers of the SBD head.
Each feature is resized and concatenated into a single tensor, again concatenated with the features obtained by the PPM.
The tensor undergoes two $1\times1$ convolutional kernels to mix the features in the channel direction.
Note that the number of convolutions can be modified.
Finally, the features are split into the original shape and concatenated to the original side layer to be upsampled and fused.

In Figure \ref{fig:twostream_sbcb_diagram}, we show explicit feature fusion by applying the two-stream architecture proposed in GSCNN.
We treat the SBD head as the Shape Stream, the final feature obtained in the Fuse Layer, and apply a $1\times1$ convolutional kernel similar to how GSCNN used the features from the Shape Stream.

\end{appendices}

\bibliographystyle{unsrtnat}
\bibliography{refs}  

\end{document}